\documentclass[sigconf]{acmart}

\usepackage{newfloat}
\usepackage{listings}
\usepackage{amsfonts}
\usepackage{stfloats}
\usepackage{amsmath}
\usepackage{booktabs}
\usepackage{multicol}
\usepackage{graphicx,subfigure}
\usepackage{multirow}
\usepackage{import}
\usepackage{algorithm}
\usepackage{algorithmic}
\usepackage{caption}
\usepackage{pifont}
\newcommand{\cmark}{\ding{51}}%
\newcommand{\xmark}{\ding{55}}%
\usepackage{hyperref}[colorlinks,linkcolor=blue]
\usepackage[english]{babel}
\newtheorem{myDef}{Definition}
\newtheorem{myAspt}{Assumption}

\usepackage{amsthm}

\AtBeginDocument{%
  \providecommand\BibTeX{{%
    \normalfont B\kern-0.5em{\scshape i\kern-0.25em b}\kern-0.8em\TeX}}}

\setcopyright{none}
\copyrightyear{2018}
\acmYear{2018}
\acmDOI{XXXXXXX.XXXXXXX}

\acmPrice{15.00}
\acmISBN{978-1-4503-XXXX-X/18/06}

\begin{document}

\title{Multi-Dimensional Ability Diagnosis for \\ Machine Learning Algorithms}

\author{Qi Liu}
\email{qiliuql@ustc.edu.cn}
\affiliation{%
  \institution{University of Science and Technology of China}
  \institution{State Key Laboratory of Cognitive Intelligence}
  \city{Hefei}
  \state{Anhui}
  \country{China}
}

\author{Zheng Gong}
\email{gz70229@mail.ustc.edu.cn}
\affiliation{%
  \institution{University of Science and Technology of China}
  \institution{State Key Laboratory of Cognitive Intelligence}
  \city{Hefei}
  \state{Anhui}
  \country{China}
}

\author{Zhenya Huang}
\email{huangzhy@ustc.edu.cn}
\affiliation{%
  \institution{University of Science and Technology of China}
  \institution{State Key Laboratory of Cognitive Intelligence}
  \city{Hefei}
  \state{Anhui}
  \country{China}
}

\author{Chuanren Liu}
\email{cliu89@utk.edu}
\affiliation{%
  \institution{Business Analytics and Statistics}
  \institution{The University of Tennessee}
  \state{Knoxville}
  \country{USA}
}

\author{Hengshu Zhu}
\email{zhuhengshu@vip.163.com}
\affiliation{%
  \institution{Career Science Lab}
  \institution{BOSS Zhipin}
  \state{Beijing}
  \country{China}
}

\author{Zhi Li}
\email{zhilizl@sz.tsinghua.edu.cn}
\affiliation{%
  \institution{Shenzhen International Graduate School}
  \institution{Tsinghua University}
  \city{Shenzhen}
  \state{Guangdong}
  \country{China}
}

\author{Enhong Chen}
\email{cheneh@ustc.edu.cn}
\affiliation{%
  \institution{University of Science and Technology of China}
  \institution{State Key Laboratory of Cognitive Intelligence}
  \city{Hefei}
  \state{Anhui}
  \country{China}
}

\author{Hui Xiong}
\email{xionghui@ust.hk}
\affiliation{%
  \institution{Hong Kong University of Science and Technology (Guangzhou)}
  \city{Guangzhou}
  \state{Guangdong}
  \country{China}
}




\begin{abstract}
  Machine learning algorithms have become ubiquitous in a number of applications (e.g. image classification). However, due to the insufficient measurement of traditional metrics (e.g. the coarse-grained \emph{Accuracy} of each classifier), substantial gaps are usually observed between the real-world performance of these algorithms and their scores in standardized evaluations. In this paper, inspired by the psychometric theories from human measurement, we propose a task-agnostic evaluation framework \emph{Camilla}, where a multi-dimensional diagnostic metric \emph{Ability} is defined for collaboratively measuring the multifaceted strength of each machine learning algorithm. Specifically, given the response logs from different algorithms to data samples, we leverage cognitive diagnosis assumptions and neural networks to learn the complex interactions among algorithms, samples and the skills (explicitly or implicitly pre-defined) of each sample. In this way, both the abilities of each algorithm on multiple skills and some of the sample factors (e.g. sample \emph{difficulty}) can be simultaneously quantified. We conduct extensive experiments with hundreds of machine learning algorithms on four public datasets, and our experimental results demonstrate that \emph{Camilla} not only can capture the pros and cons of each algorithm more precisely, but also outperforms state-of-the-art baselines on the metric reliability, rank consistency and rank stability.
\end{abstract}

\ccsdesc[300]{Computing methodologies~Statistical relational learning}
\ccsdesc[100]{Applied computing~Collaborative learning}

\keywords{Machine Learning; Cognitive Diagnosis; Human Measurement}



\maketitle

\section*{Introduction}
 
Recent years have witnessed the astounding strides of machine learning algorithms in various fields, including computer vision, natural language processing, data mining and so on~\cite{lecun2015deep,DBLP:conf/acl/00020ZZ020,becht2021high,leist2022mapping,jablonka2023machine}. For evaluating the performance of these algorithms, different metrics have been adopted, such as the value of \emph{Accuracy} or \emph{F1-score} in classification tasks. Given these metrics, some of machine learning algorithms can even outperform humans according to the standardized evaluation of public leaderboards~\cite{imagenet, liang2020xglue, DBLP:conf/nips/HuFZDRLCL20}.







\begin{figure*}[t]
    \centering
    \includegraphics[width=1\textwidth]{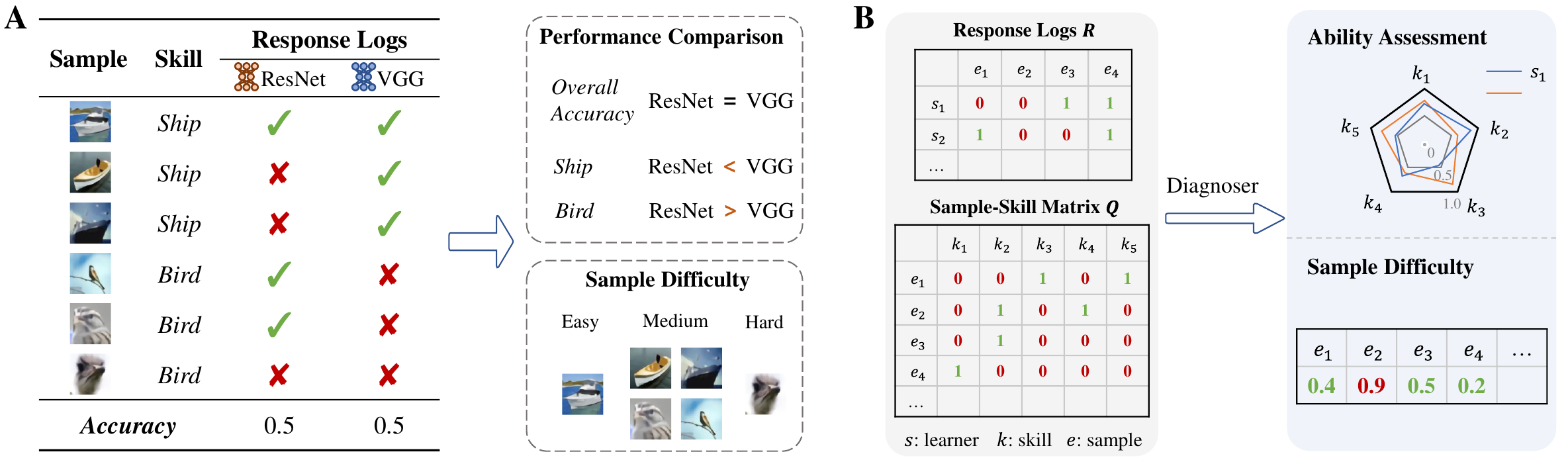}
\centering
\caption{\label{Fig.introduction}\textbf{A}, a toy example for understanding the performance of machine learning algorithms. From the left response logs, we can intuitively infer the difference in both algorithms' performance~(the right top part) and the  difficulty of data samples~(the right bottom part). \textbf{B}, the formal procedure of cognitive diagnosis task. Given the learner-sample response matrix $R$ and the sample-skill relevancy matrix $Q$, the cognitive diagnosis task aims to train a diagnoser which can assess the \emph{Ability} of different learners $s$~(e.g. ResNet and VGG in Figure 1a) on different skills $k$~(e.g. \emph{Ship} and \emph{Bird} in Figure 1a) and quantify the factors~(e.g. \emph{difficulty}) of samples $e$~(e.g. the images in Figure 1a) collaboratively.}
\end{figure*}

 Unfortunately, most of traditional metrics are insufficient in capturing the pros and cons of each machine learning algorithm~\cite{drummond2010warning,gosiewska2022interpretable, orzechowski2022generative}. Substantial gaps are usually observed between the high evaluation scores of these algorithms in standardized settings and the low performance of them in real-world scenarios~\cite{geirhos2020shortcut,voudouris2022direct}. One of the major reasons is that these metrics usually just output a single numerical value, which not only is too coarse-grained for representing the multifaceted ability of each algorithm but also cannot consider the wide difference of data samples~\cite{ethayarajh2022understanding,hernandez2021general}. As the toy example shown in Figure~\ref{Fig.introduction}A, two machine learning algorithms (i.e. ResNet and VGG) for image classification have just been evaluated on six data samples, where different skills are required for algorithms correctly responding to~(predicting) each sample\footnote{Without loss of generality, here we use data labels to represent skills.}. Though the overall \emph{Accuracy} reached by ResNet and VGG are the same~(i.e. 0.5), the performance of these two algorithms on samples with different skills is completely different. For instance, VGG may outperform ResNet on the samples of \emph{Ship}, while performs poorly on \emph{Bird}. Meanwhile, we can infer that the difficulty of responding to each sample correctly is also quite different, and one step further, these samples should have different contributions for evaluating the strength of each algorithm. Therefore, without the help of such fine-grained and multifaceted analysis, just devoting the efforts to improve the overall scores of machine learning algorithms on traditional metrics like \emph{Accuracy} cannot lead to the satisfactory performance of these algorithms in practical applications.

As is well known, a significant proportion of noticeable improvement in machine learning architectures actually benefits from the consistent inspiration of the way human learning~\cite{geirhos2020shortcut,udrescu2020ai}. For instance, both \textit{Curriculum learning}~\cite{bengio2009curriculum, graves2017automated} and \textit{self-paced learning}~\cite{kumar2010self, jiang2014self} are inspired by highly organized human education systems, i.e. training the algorithms with easy samples first and gradually transforming to the hard examples can contribute to faster convergence and lower generalization error. Similarly, the evaluation of machine learning algorithms may also benefit from the more comprehensive and fine-grained measurement of human learning. To this end, in this paper, inspired by the psychometric theories from human measurement~\cite{newell1972human,nichols2012cognitively}, we propose a general  \emph{\textbf{C}}ognitive di\emph{\textbf{a}}gnosis framework for \emph{\textbf{m}}ach\emph{\textbf{i}}ne \emph{\textbf{l}}earning algorithm eva\emph{\textbf{l}}u\emph{\textbf{a}}tion~(\emph{Camilla}). Under this framework, a multi-dimensional diagnostic metric \emph{Ability} is defined for collaboratively measuring the multifaceted strength of each algorithm. Specifically, given the response logs from different well-trained algorithms to data samples, we leverage cognitive diagnosis assumptions~\cite{wang2020neural} and neural networks to learn the complex interactions among algorithms, samples and the required skills for algorithms correctly responding to each sample. Here, the skills can be explicitly given (e.g. the skills shown in Figure~\ref{Fig.introduction}A) or implicitly pre-defined as latent factors. In this way, both the abilities of each algorithm on multiple skills and some of the sample factors (e.g. sample \emph{difficulty}) can be simultaneously quantified. Finally, we conduct extensive experiments with hundreds of machine learning algorithms on four public datasets of classification and regression tasks. The results demonstrate that our evaluation can capture the pros and cons of each algorithm. Moreover, based on the diagnosed algorithm abilities, future response behaviors of each machine learning algorithm on unknown data samples (e.g. the correct/wrong classifications of one classifier) can be precisely predicted in an interpretable way.

 To the best of our knowledge, this is the first comprehensive attempt for measuring the multifaceted strength of each machine learning algorithm through exploring the connections between the research on psychometric theories and machine learning evaluation. The proposed solution can be generally applied in broad applications for evaluating different types of machine learning algorithms more comprehensively, improving the quality assessment of data samples~\cite{braunstein2021expectation,monsalve2022analysis}, and will also be helpful for training/testing algorithms more efficiently and so on. 
 

\section*{Preliminary and Problem Definition}






In this section, we introduce the basic assumptions and the definition of the cognitive diagnosis problem for machine learning algorithms.
For ease of understanding, we call the machine learning algorithms to be evaluated as \textbf{learners}, and the cognitive diagnosis method as \textbf{diagnoser}. Our goal in developing the diagnoser is to estimate the learners' multidimensional \emph{Ability} on specific \emph{skills}, which are defined as follows:
\begin{myDef}\label{def:ability}
\emph{Ability} is a multidimensional metric for quantifying the proficiency of learner on specific skills. Each entry (with the value in the range of  {\rm [0, 1]}) represents one skill, and the skill can be explicitly given or pre-defined as latent factors.
\end{myDef}

As shown in Figure~\ref{Fig.introduction}A, each learner after training may perform well on some of the samples (or skills) while perform poorly on others.
Similarly, one learner may outperform another learner on a part of samples (or skills) whereas fail on the other samples.
Therefore, we propose a multi-dimensional diagnostic metric \emph{Ability} to quantify such internal and external performance differences.
We summarize such observations in Assumption 1:

\begin{myAspt}\label{Ass:different}
The \emph{Ability} of learners on different skills can be different, and the required skills for learners correctly responding to each sample can also be different.
\end{myAspt}

We aim to design the diagnoser \emph{Camilla} that can compute the multi-dimensional \emph{Ability} of each well-trained learner by mining the complex interactions among learners, samples and skills.
For instance, the \emph{Ability} of ResNet on \emph{Ship} and \emph{Bird} in Figure~\ref{Fig.introduction}A may be 0.2 and 0.8, respectively.
Note that, our \emph{Ability} is not simply a skill-specific version of the traditional metrics like \emph{Accuracy}, because we also collaboratively consider the learning \emph{difficulty} and \emph{discrimination} of different data samples.
Here, the \emph{discrimination} indicates the capability of samples to differentiate the proficiencies of learners.
Finally, \emph{Ability} needs to follow the monotonicity assumption~\cite{rosenbaum1984testing, wang2020neural}:

\begin{myAspt}\label{Ass:Mono}
The probability of correct response to the sample is monotonically increasing with the \emph{Ability} of learners.
Similarly, giving different samples, this probability for one learner is monotonically decreasing with the sample \emph{difficulty}.
\end{myAspt}

To design our task-agnostic diagnoser, we consider a well-trained learner set $S = \{s_1, ..., s_N\}$ and a data sample set $E = \{e_1, ..., e_M\}$. After running each learner $s_i$ on $E$ we can get response logs $R$, denoted as a set of triple $(s_i,e_j,r_{ij}), s_i\in S, e_j\in E$ and $r_{ij}$ is the response score. Specifically, in the case of classification tasks, $r_{ij}=1$ if learner $s_i$ answers the class label of sample $e_j$ correctly and $r_{ij}=0$ otherwise. While for regression tasks, we obtain the response scores through two steps. First, we adopt sample-wise min-max normalization to rescale the absolute errors between the learner's answer of each sample and the corresponding sample label to the range from 0 to 1. Then, since the absolute errors are monotonically decreasing with the \emph{Ability} of learners, we reverse the normalized absolute errors as response scores by using 1 minus them. Meanwhile, an explicitly or implicitly pre-defined sample-skill relevancy matrix $Q$ should also be given. $Q = \{Q_{ij}\}_{M\times K}$, where $K$ is the number of skills, $Q_{ij}=1$ denotes the sample $e_i$ is 100\% related to the skill $k_j$ and $Q_{ij}=0$ otherwise.
Formally, the cognitive diagnosis problem for machine learning can be defined as (a toy example is shown in Figure~\ref{Fig.introduction}B):

\begin{myDef}\label{def:diagnoser}
Given the learner-sample response matrix $R$ and the sample-skill relevancy matrix $Q$, the cognitive diagnosis task aims to train a diagnoser which can assess the \emph{Ability} of different learners on different skills.
\end{myDef}

With the help of diagnoser, not only the \emph{Ability} of each learner on specific skills can be quantified but also some of the sample factors (e.g. sample \emph{difficulty}) are collaboratively quantified.
Before introducing our diagnoser \emph{Camilla}, we should note that the performance of a diagnoser is difficult to evaluate as we cannot obtain the ground-truth proficiency of learners.
Therefore, we will investigate the reliability of diagnosers indirectly through their performance in predicting the response of learners on unknown samples. Let's take the classification task as an example for illustration, there are totally two cases for a successful prediction of the diagnoser: (1) the diagnoser predicts that one learner can answer the class label of this sample correctly, and the learner does answer it correctly; (2) the diagnoser predicts that one learner can not answer the class label correctly, and the learner does give a wrong answer.







\subsection*{\emph{Camilla}: Task-Agnostic Cognitive Diagnostic Framework for Machine Learning Evaluation}


\begin{figure*}[t]
  \centering
  \includegraphics[width=1\textwidth]{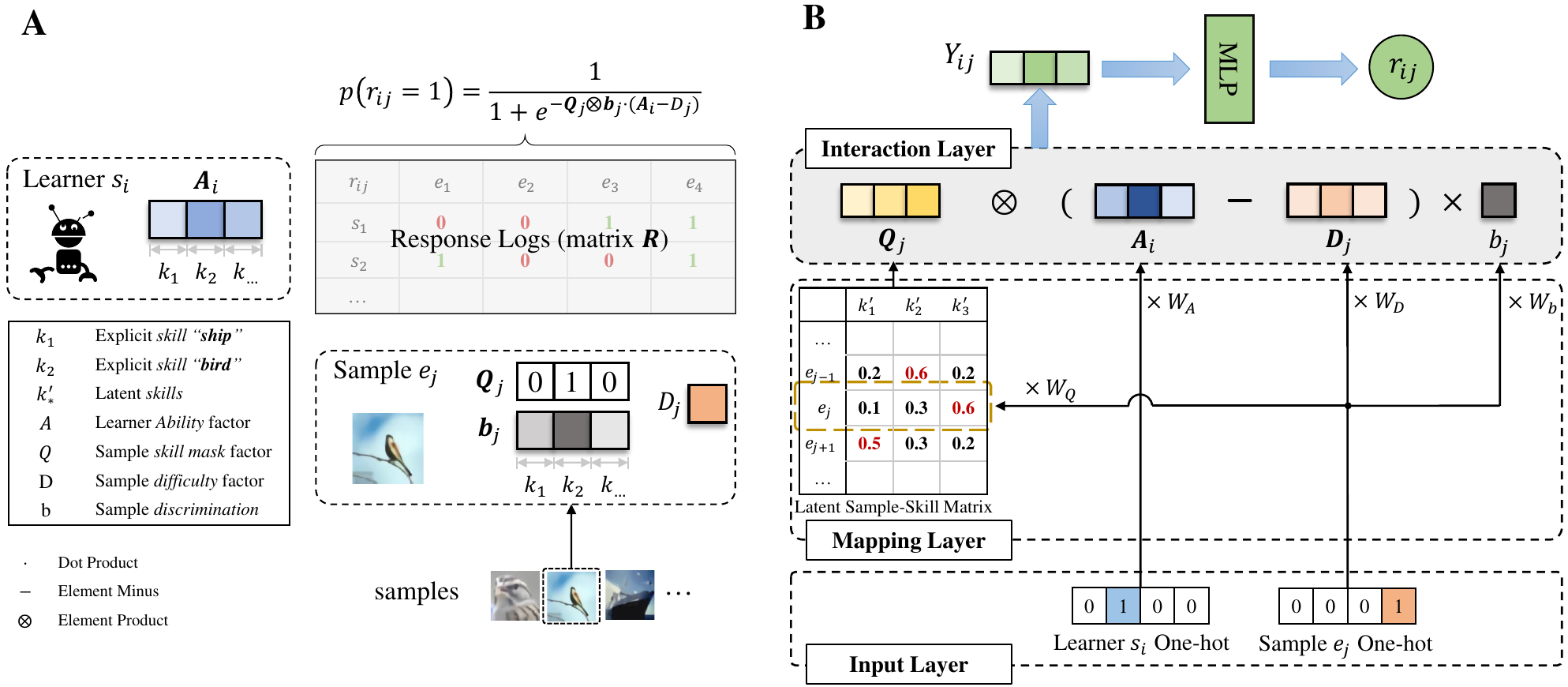}
\centering
\caption{\textbf{The overview structures of our proposed two diagnosers \emph{Camilla-Base} and \emph{Camilla}.} \textbf{A}, the architecture of \emph{Camilla-Base}, which extends the multidimensional \emph{Ability} of classic MIRT to the proficiency of learners on explicit skills by incorporating sample \emph{skill mask} factor $Q_j$.  \textbf{B}, \emph{Camilla} characterizes the sample \emph{skill mask} factor $Q_j$ as a learnable factor and models the interaction between learners and samples via an interaction layer along with a multi-layer perception.}
\label{Fig.camilla}
\end{figure*}

In this section, we introduce the schematic architectures of our proposed diagnostic framework \emph{Camilla} and its basic version \emph{Camilla-Base}.
The most notable difference between \emph{Camilla-Base} and \emph{Camilla} is that \emph{Camilla} can adapt to the machine learning tasks where the explicit sample-skill matrix is unknown and we have to characterize the sample \emph{skill mask} factor as a learnable factor. Here, \emph{skill mask} depicts the skills that are most needed for correct response of the sample by masking the other unrelated skills. Without loss of generality, in the following, we use \emph{Camilla}s to represent both \emph{Camilla-Base} and \emph{Camilla}. Inspired by existing cognitive diagnosis models~\cite{dibello200631a, wang2020neural} in psychometric theories, one diagnoser usually includes three components: (1) the input representations of learners and samples, (2) the factors of learners and samples to be measured and (3) the interactions among them. Therefore, given the response logs in matrix $R$, where each entry $r_{ij}$ denotes the response score from learner $s_i$ to sample $e_j$, we design \emph{Camilla}s by taking the one-hot sample representation and one-hot learner representation as input. Then we characterize the learner and the sample with the learner's diagnostic factor (i.e. \emph{Ability}) and the sample's diagnostic factors (e.g. \emph{difficulty}). Finally, by taking into account the factors of learners and samples interactively, the target of \emph{Camilla}s is to predict the ground truth response of the learner to each sample (i.e. $r_{ij}$ in $R$).



Specifically, inspired by the psychometric theories from human measurement~\cite{newell1972human,nichols2012cognitively}, we propose a general diagnostic framework \emph{Camilla} for evaluating machine learning algorithms. For better illustration, we start with \emph{Camilla-Base}, a basic  version of \emph{Camilla} which uses classic MIRT~\cite{reckase2009multidimensional} as the backbone and incorporates explicit sample-skill mapping~\cite{tatsuoka1995architecture} to measure the \emph{Ability} of learners on explicit skills. Then, we augment this basic diagnoser with latent sample \emph{skill mask} factors and a multi-layer perception~(MLP) to accommodate most various machine learning tasks where the sample-skill matrix is unknown, without loss of precision and interpretability for capturing the interactions between learners and samples. Figure~\ref{Fig.camilla}A and Figure~\ref{Fig.camilla}B illustrate the schematic overview of \emph{Camilla-Base} and \emph{Camilla}. In the following of this section, we will introduce the workflows, training target, computation of the overall ability value and the implementation details of \emph{Camilla}s, respectively.


\noindent\textbf{Camilla-Base.}~\emph{Input.} The input of \emph{Camilla-Base} includes the response logs/matrix $R$ from different well-trained learners to data samples, a pre-defined sample-skill matrix $Q$, and the representations of learners and samples.

\noindent 1)~\emph{Learner representation.} In this work, we focus on the \emph{Ability} assessment of individual learners. Therefore, we represent learner $s_i$ input as one-hot vector $s_i \in \{0, 1\}^{1\times N}$.



\noindent 2)~\emph{Sample representation.} Similar to learner representation, we use one-hot vector $e_j \in \{0, 1\}^{1\times M}$ to represent sample $e_j$.

\noindent\emph{Mapping layer.} The mapping layer consists of factors that characterize the latent trait of learners and samples.


\noindent 1)~\emph{Learner Ability.} The meaning of multidimensional \emph{Ability} factor is given in Definition~\ref{def:ability}. Each dimension of \emph{Ability} corresponds to the proficiency of learners on a specific skill. We characterize the \emph{Ability} factor $A_i$ of learner $s_i$ as follows:
\begin{equation}
    A_i = s_i  \times  W_{A},
\end{equation}
where $A_i\in \mathbb{R}^{1\times K}$, $W_{A} \in \mathbb{R}^{N \times K}$ is a trainable transformation matrix, $K$ is the number of explicit skills pre-defined in the sample-skill matrix $Q$. During implementations, we can fix each value of $A_i$ into the range of $[0, 1]$, e.g. via Sigmoid function. 


\noindent 2)~\emph{Sample factor.} The sample factors characterize the latent trait of samples. In this paper, we assume the sample factors consist of three components: \emph{skill mask} factor, \emph{difficulty} factor and \emph{discrimination} factor.


\noindent (1)~\emph{Sample skill mask factor.} As the proficiency of learners are characterized by a multi-dimensional \emph{Ability}, we consider that different samples correspond to different dimensions of the learner \emph{Ability} factor, which can be evidenced by the sample-skill matrix $Q$. Specifically, the \emph{skill mask} factor depicts the skills that are most needed for a correct response of the sample by masking the other unrelated skills. The \emph{skill mask} factor $Q_j$ of sample $e_j$ is fixed as:
\begin{equation}
     Q_j = e_j  \times  Q,
\end{equation}
where $Q_j\in \{0, 1\}^{1\times K}$, $e_j$ denotes the one-hot vector of sample $e_j$. 


\noindent (2)~\emph{Sample difficulty factor.} Similar to curriculum learning~\cite{bengio2009curriculum} and self-paced learning~\cite{kumar2010self}, which suggest that there are differences between the difficulty of samples, we also argue that the \emph{difficulty} factor of samples should be taken into account in the learner \emph{Ability} assessment. Specifically, if one learner can correctly respond to a data sample while other learners cannot, then this sample with high \emph{difficulty} should contribute more to this learner's corresponding \emph{Ability}, and vice versa. Therefore, capturing the relationship between the \emph{Ability} factor of learners and the \emph{difficulty} factor of samples can help to more precisely predict how sure the learner can respond to the sample correctly. Following the classic MIRT\cite{reckase2009multidimensional}, the \emph{difficulty} factor of sample $e_j$ is represented as:
\begin{equation}
    D_j = e_j  \times  W_{D},
\end{equation}
where $D_j\in \mathbb{R}$, $e_j$ is the one-hot representation of sample $e_j$, and $W_D$ is a trainable transformation matrix.

\noindent (3)~\emph{Sample discrimination factor.} Following the psychometrics~\cite{lord1952theory}, sample \emph{discrimination} factor indicates the capability of sample $e_j$ to differentiate the mastery degree of learners. We represent the sample \emph{discrimination} factor as:
\begin{equation}
    b_j = e_j  \times  W_{b},
\end{equation}
where $b_j\in \mathbb{R}^{1\times K}$, and $W_{b}  \in \mathbb{R}^{M \times K}$ is a trainable transformation matrix.

\noindent\emph{Interaction layer.} After giving the input and mapping representation, the way to define a function for modeling the complex interactions among learners, samples and skills is one of the most important components for a diagnoser. Inspired by MIRT~\cite{reckase2009multidimensional}, we adopt an interaction layer which  outputs the probability $r_{ij}\in R$  that the learner $s_i$ responds to the sample $e_j$ correctly via the comparison between the multi-dimensional learner \emph{Ability} $A_i$ and the \emph{difficulty} $D_j$ of the sample in covered skills. In this layer, we define the diagnose function as (i.e. Eq.(\ref{equ:Camilla-Base})):
\begin{equation}
    \label{equ:Camilla-Base}
    p\left(r_{i j}=1|A_i, Q_j, b_j, D_j\right)=\frac{1}{1+e^{-Q_j \otimes b_j \cdot\left(A_i-D_{j}\right)}},
\end{equation}
where $\otimes$ is element-product and $\cdot$ is dot-product.

\noindent\textbf{Camilla.}~\emph{Input}.~For general applications in various machine learning tasks where the sample-skill matrix $Q$ may not be available, \emph{Camilla} takes only the response logs/matrix $R$ and the representations of learners and samples as input, and the definitions of learner and sample representations are the same as \emph{Camilla-Base}.

\noindent\emph{Mapping layer.} In the mapping layer of \emph{Camilla}, the principles of learner and sample factors inherit from \emph{Camilla-Base}. However, \emph{Camilla} utilizes a multi-layer perception (MLP), instead of the simplified
equation Eq.(\ref{equ:Camilla-Base}) in \emph{Camilla-Base}, to more precisely capture the interactions between learners and samples, and we revise the implementation of each factor to accommodate the general applications of \emph{Camilla}.

\noindent 1)~\emph{Learner Ability.}~To enhance the capability of MLP, the \emph{Ability} $A_i^{'}$ of learner $s_i$ are rescaled to [0, 1] via Sigmoid function:
\begin{equation}
    A_i^{'} = \text{Sigmoid}(s_i  \times  W_{A}^{'}),
\end{equation}
where $A_i^{'}\in (0, 1)^{1\times K^{'}}$, $W_{A}^{'} \in \mathbb{R}^{N \times K^{'}}$, and the dimension $K^{'}$ is a hyperparameter. 

\noindent 2)~\emph{Sample factors.}~Since there are various machine learning tasks where the sample-skill matrix $Q$ may be not explicitly accessible, we characterize the \emph{skill mask} factor $Q_j^{'}$ as a learnable factor for the representation of sample $e_j$:
\begin{equation}
    Q_j^{'} = \text{Softmax}(e_j  \times  W_{Q}^{'}),
\end{equation}
where $Q_j^{'}\in (0, 1)^{1\times K^{'}}$, $e_j$ denotes the one-hot vector of sample $e_j$, and $W_{Q}^{'}$ is a trainable
transformation matrix. The value of $Q_j^{'}$ in each dimension indicates the degree to which the sample $e_j$ is associated with each latent skill. In order to more precisely diagnose multidimensional \emph{Ability}, we characterize the latent traits of samples with multidimensional \emph{difficulty} factor $D_j^{'} \in (0, 1)^{1 \times K^{'}}$ and unidimensional \emph{discrimination} factor $b_j^{'}\in (0, 1)$. $D_j^{'}$ indicates the sample difficulty in each latent skill and has a one-to-one correspondence with the multifaced learner \emph{Ability} $A_i^{'}$. We obtain them by:
\begin{equation}
    D_j^{'} = \text{Sigmoid}(e_j  \times  W_{D}^{'})~~\text{and}~~b_j^{'} = \text{Sigmoid}(e_j  \times  W_{b}^{'}),
\end{equation}
where $W_{D}^{'} \in \mathbb{R}^{M \times K^{'}}$ and $W_{b}^{'} \in \mathbb{R}^{M \times 1}$ are trainable.

\noindent\emph{Interaction layer.} After giving the input and mapping representation, the way to define a function for modeling the complex interactions among learners, samples and the skills is one of the most important components for a diagnoser. Inspired by the cognitive diagnosis models in psychometric
theories~\cite{lord1952theory,wang2020neural}, we adopt an interaction layer which can combine cognitive diagnosis assumptions and neural networks for keeping both effectiveness and interpretability in machine learning evaluation. Specifically, the interaction layer outputs the probability $r_{ij} \in R$  that the learner $s_i$ responds to the sample $e_j$ correctly via the comparison between the multi-dimensional learner \emph{Ability} $A_i^{'}$ and the \emph{difficulty} $D_j^{'}$ of the sample in covered skills. In this layer, we first define the diagnose function as:
\begin{equation}\label{equ:Camilla}
     {Y}_{ij} = Q_j^{'} \otimes(A_i^{'} - D_j^{'}) \times b_j^{'},
\end{equation}
where $Y_{ij} \in  \mathbb{R}^{1\times K^{'}}$, $\otimes$ is element-product and $b_j$ is sample \emph{discrimination}. Then, we aggregate $Y_{ij}$ with a fully connected multi-layer perception (MLP) as follows:
\begin{equation}
\label{interaction_layer}
o = \text{sigmoid}(W_1Y_{ij} + b_1),~~
r_{ij} = \text{sigmoid}(W_2o + b_2).
\end{equation}

Following the solution of NeuralCD\cite{wang2020neural} in human performance
measurement, we also promise
the monotonicity in Assumption \ref{Ass:Mono} by normalizing the MLP where the parameters are restricted to be non-negative.

\noindent\textbf{Diagnoser training.}
The loss function of diagnosers depends on the specific type of machine learning task. For instance, the cross entropy between prediction probability $\hat{r}_{ij}$ and the ground truth response of the learner $r_{ij}$ can serve as the loss function $\mathcal{L}_{\text{cls}}$ for classification learners\footnote{For instance, $r_{ij}=1$ represents the learner classifies the sample correctly (no matter the class labels of the sample), and 0 otherwise.}. As for regression tasks, we select the mean squared error as loss function $\mathcal{L}_{\text{reg}}$.
\begin{equation}
    \begin{split}
    \mathcal{L}_{\text{cls}}&=-\frac{1}{N}\frac{1}{M}\sum\limits_{i=1}^N\sum\limits_{j=1}^M\left(\hat{r}_{ij} \log r_{ij}+\left(1-\hat{r}_{ij}\right) \log \left(1-r_{ij}\right)\right), \\
    \mathcal{L}_{\text{reg}}&=\frac{1}{N}\frac{1}{M}\sum\limits_{i=1}^N\sum\limits_{j=1}^M \left(\hat{r}_{ij} - r_{ij}\right)^2.
    \end{split}
\end{equation}

With this loss function and the observed input data (e.g. $R$), both the multi-dimensional \emph{Ability} of learners and the factors (e.g. \emph{difficulty} and \emph{discrimination}) of samples can be estimated after inferring \emph{Camillas}. 


\begin{figure*}[t]
    \centering
    \includegraphics[width=0.7\textwidth]{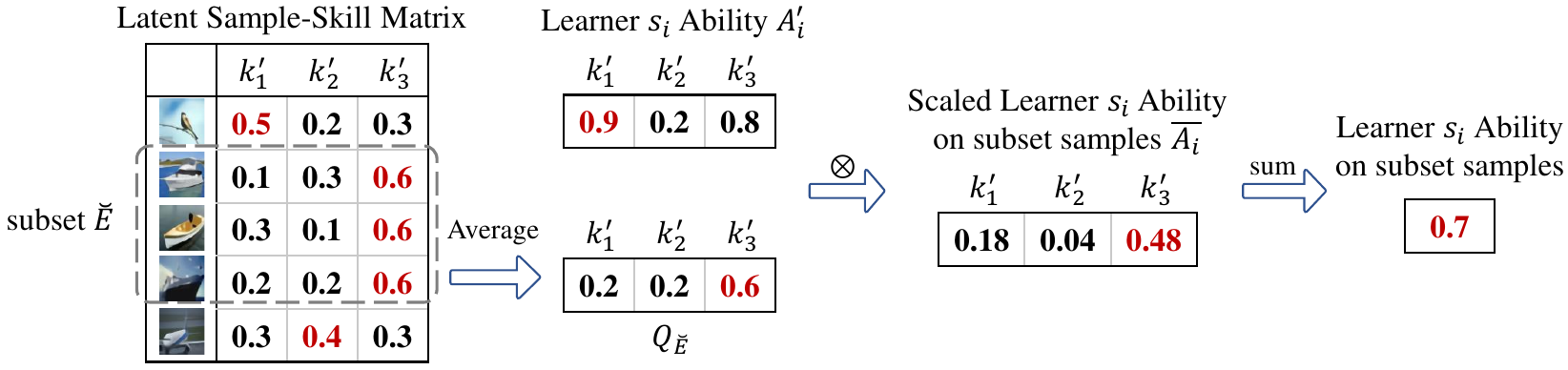}
    \caption{An example for the way of getting the specific proficiency from the learner to a subset of samples with \emph{Camilla}.}
    \label{Fig.compare}
\end{figure*}

\noindent\textbf{Computation of the overall ability value for each learner.}~In the foregoing, we show the way to output the multi-dimensional \emph{Ability} for each learner based on \emph{Camilla-Base} and \emph{Camilla}, respectively. Considering that it is also necessary to get the overall ability in some applications (e.g. Figure~\ref{Fig.COR}), we then give the solution for summarizing a unidimensional ability value of each learner from this \emph{Ability} vector.

\noindent\emph{Camilla-Base.}~\emph{Camilla-Base} exploits the explicit sample-skill matrix to get the \emph{skill mask} factors of data sample, and the multi-dimensional \emph{Ability} $A_i$ of learner $s_i$ will be output after training the diagnoser (i.e. fitting Eq.(\ref{equ:Camilla-Base})) , where each dimension of $A_i$ indicates the learner proficiency on specific skill. Then, we can obtain the overall \emph{Ability} $a_i$ of $s_i$ on dataset $E$ by
\begin{equation}
    a_i = \frac{1}{|E|}\sum\nolimits_{e_j\in E} \frac{Q_j \cdot A_i}{\sum\nolimits_{k=1}^K Q_{jk}} ,
\end{equation}
where $a_i \in [0, 1]$, $A_i \in \mathbb{R}^{1\times K}$, and $\sum\nolimits_{k=1}^K Q_{jk}$ is the number of explicit skills of sample $e_j$.


\noindent\emph{Camilla.}~After training \emph{Camilla} (e.g. fitting Eq.(\ref{equ:Camilla})), the vector of learner \emph{Ability} $A_i^{'}\in [0, 1]^{1\times K^{'}}$ is output as the assessment result of machine learning algorithm $s_i$ on multiple skills. This original $A_i$ can also help us estimate the specific proficiency from $s_i$ to any single or subset of samples even if the sample-skill matrix $Q$ is implicitly defined as latent factors (i.e. both the meaning of each skill and the sample-skill relation is unknown). As shown in Figure~\ref{Fig.compare}, to output the learner $s_i$'s ability $a_{\breve{E}}^i$ on a specific subset $\breve{E} \subseteq E$ of data samples, we first derive the overall \emph{skill mask} factor of the subset $Q_{\breve{E}}$ by averaging the \emph{skill mask} factors of the samples in this subset: 
\begin{equation}
    Q_{\breve{E}} = \frac{1}{|\breve{E}|}\sum\nolimits_{e_j\in \breve{E}} Q_j^{'}. 
\end{equation}
Then we summarize the overall ability ($a_{\breve{E}}^i$) of learner $s_i$ on the subset $\breve{E}$ of samples by summing the learner \emph{Ability} $A_i^{'}$ weighted by the \emph{skill mask} factor $Q_{\breve{E}}$:
\begin{equation}
a_{\breve{E}}^i = \sum\nolimits_{k=1}^{K^{'}}\overline{A_{ik}}, \quad \text{and} \quad
    \overline{A_i} = Q_{\breve{E}} \otimes A_i^{'},
\end{equation}
where $\otimes$ is element-product, and $\overline{A_i}$ is the learner \emph{Ability} scaled by $Q_{\breve{E}}$.


Please note that the \emph{difficulty} factor $D_j^{'}$ of each sample $e_j$ in \emph{Camilla} is also a vector, and we can get the overall and unidimensional \emph{difficulty} of any sample similar to the way of computing $a_{\breve{E}}^i$, i.e. replacing $A_i^{'}$ by $D_j^{'}$ in Figure~\ref{Fig.compare}.

\section*{Evaluation}

\textbf{Dataset Description.} We evaluate our proposed framework \emph{Camilla} and the metric \textit{Ability} on two real-world classification datasets - Titanic\footnote{\href{https://www.kaggle.com/c/titanic}{https://www.kaggle.com/c/titanic}} and CIFAR-100~\cite{cifar} as well as two widely used regression datasets - ESOL\cite{delaney2004esol} and Diamond\footnote{\href{https://www.kaggle.com/shivam2503/diamonds}{https://www.kaggle.com/shivam2503/diamonds}}.

Regarding the classification tasks, Titanic is a data mining competition containing passenger and crew data in the ``unsinkable" Titanic shipwrecks with the dichotomous aim of predicting what sorts of samples (i.e. passengers and crews) are more likely to survive. As for this data, we implement $353$ mainstream machine learning algorithms as learners including k-Nearest Neighbors~\cite{fix1989discriminatory}, decision tree~\cite{quinlan1987simplifying} and support vector machine~\cite{cortes1995support} based on the public code platform of Kaggle. CIFAR-100 is a large image classification dataset and contains the same amount of images~(samples) in $100$ classes.
We reimplement $42$ popular algorithms with top performance as learners including ResNet~\cite{he2016deep}, VGG~\cite{DBLP:journals/corr/SimonyanZ14a} and DenseNet~\cite{huang2017densely} based on Pytorch~\cite{paszke2019pytorch}. 

In the case of regression tasks, ESOL is a standard regression dataset providing graph structures of 1,127 compounds as input and aims to predict water solubility property of them. Please note that there are no explicit skills pre-defined for graph structures. We implement 8 classic algorithms~(e.g. Random Forest and k-Neighbor Regressor) built on the fingerprints of molecules and 14 popular graph neural networks~(e.g. graph attention network~\cite{velivckovic2018graph} and graph convolution network~\cite{DBLP:conf/iclr/KipfW17}). Besides, Diamond is a large dataset for estimating the prices of diamonds through attributes~(e.g. color and clarity). We implement 157 algorithms as learners including traditional methods~(e.g. Random Forest) and ensemble learning models such as Lightgbm~\cite{ke2017lightgbm}. 

Supplementary Table~\ref{Table: statistics} summarizes the statistics of the datasets. As for the training of the learners and the generation procedure of learners' response to samples (i.e. response matrix $R$), please also refer to Supplement for detailed descriptions.

\begin{figure*}[t]
  \centering 
  \includegraphics[width=0.8\textwidth]{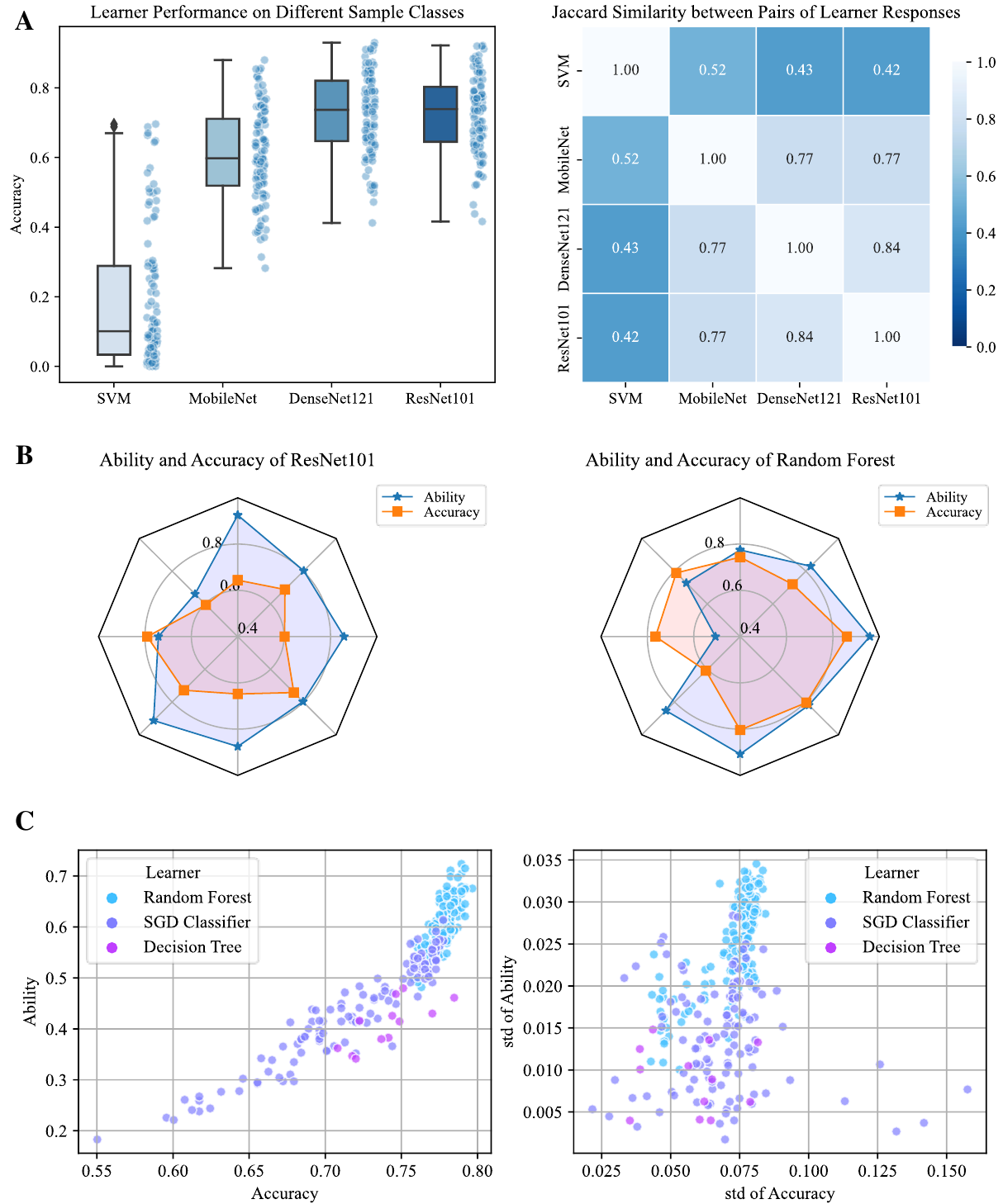}
  \caption{\textbf{Assumption validation and the comparison between Accuracy and learner \textit{Ability} diagnosed by Camilla-Base.} \textbf{A}, different performance of each learner on different sample classes of CIFAR-100~(Left Figure). \emph{Jaccard} similarity of two learners' responses in CIFAR-100~(Right Figure). \textbf{B}, \textit{Ability} assessment and classification \emph{Accuracy} of each learner on different skills~. \textbf{C}, visualization of the \emph{Ability} and \emph{Accuracy} values (and standard deviation~(std) of these values) of three kinds of learners with different hyperparameters in Titanic.}

  \label{Fig. visualization1}
\end{figure*}

\noindent\textbf{Assumption Validation.}~Supplementary Table~\ref{Table. titanic learner}, \ref{Table. CIFAR100 learner}, \ref{Table. ESOL learner} and \ref{Table. diamond learner} present the hyperparameters and overall  response performances of the learners in these datasets, respectively. From these overall performances that are measured by traditional metrics (e.g. the mean \emph{Accuracy} of each classifier and the \emph{MAE/RMSE} of each regression learner), we discover that a large number of learners get very close evaluation values. For instance, though the \emph{Accuracy} of all learners in CIFAR-100 (Supplementary Table~\ref{Table. CIFAR100 learner}) range from $0.19$ to $0.78$, the top 50\% learners' \emph{Accuracy} locate in a small range of [$0.75$, $0.78$]. Therefore, it is obviously difficult to capture the significant difference between these learners with coarse-grained metrics. To verify our assumption that the proficiency of learners on different skills are different~(Assumption \ref{Ass:different}), we compare the performance of four individual learners on different sample classes in the dataset of CIFAR-100 in Figure~\ref{Fig. visualization1}A, where we simply view each class as a skill in matrix $Q$. Specifically, in the left boxplot of Figure~\ref{Fig. visualization1}A, the average performance of ResNet101 is close to DenseNet121, while outperforms SVM a large margin~(the overall performance of learners in CIFAR-100 is summarized in Supplementary Table \ref{Table. CIFAR100 learner}). However, the \emph{Accuracy} on different sample classes of each single learner is quite different, e.g. varies from 0.4 to 0.9 for ResNet101, and this proves the different proficiency of learners on different skills. One step further, in the right heat map of Figure~\ref{Fig. visualization1}A, we obtain the \emph{Jaccard} similarity of the pair learners' responses, by counting the samples two learners both predict correctly or wrongly and then dividing by the number of all data samples. We can see the noteworthy distinction between each two learners, even for the two convolution neural networks-based learners ResNet101 and DenseNet121. This indicates that even though learners are similar in both architectures and overall \emph{Accuracy}, there is still significant difference in their fine-grained performance.

\subsection*{Interpretation of Diagnostic Factors from Case Studies}
\begin{figure*}[t]
  \centering 
  \includegraphics[width=0.75\textwidth]{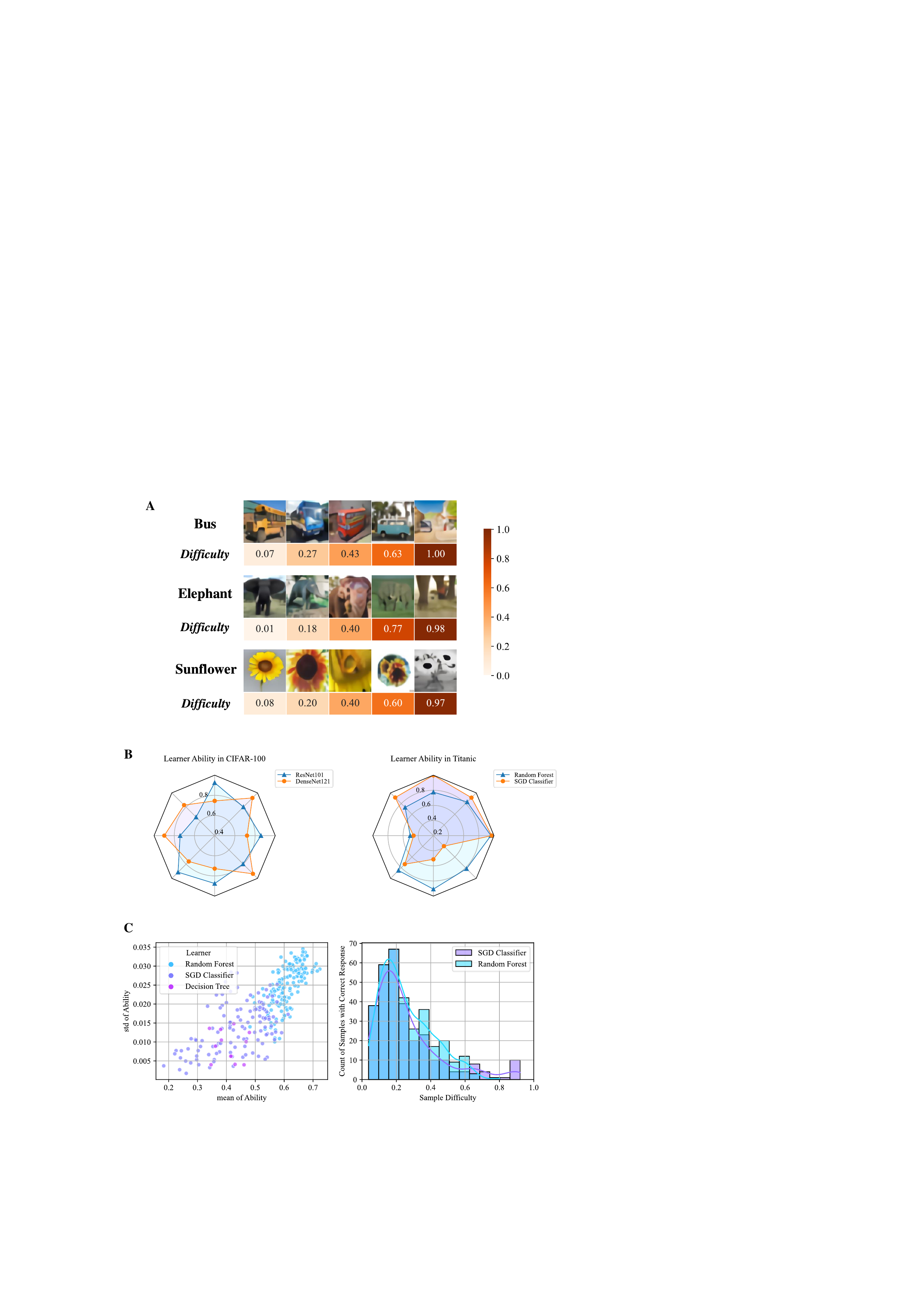}
  \caption{\textbf{Visualization of sample \textit{difficulty} and learner \textit{Ability} diagnosed by Camilla-Base.} \textbf{A}, the samples randomly selected from ``bus'', ``elephant'' and ``sunflower'' classes, and the corresponding \emph{difficulty} factors. \textbf{B}, \emph{Ability} comparison of different learners in CIFAR-100 and Titanic. \textbf{C}, mean and standard deviation of \emph{Ability} of different learners in Titanic~(Left Figure) and \emph{difficulty} distribution of the samples that different learners answer correctly~(Right Figure).}

  \label{Fig. visualization2}
\end{figure*}

\begin{figure*}[t]
  \centering 
  \includegraphics[width=1\textwidth]{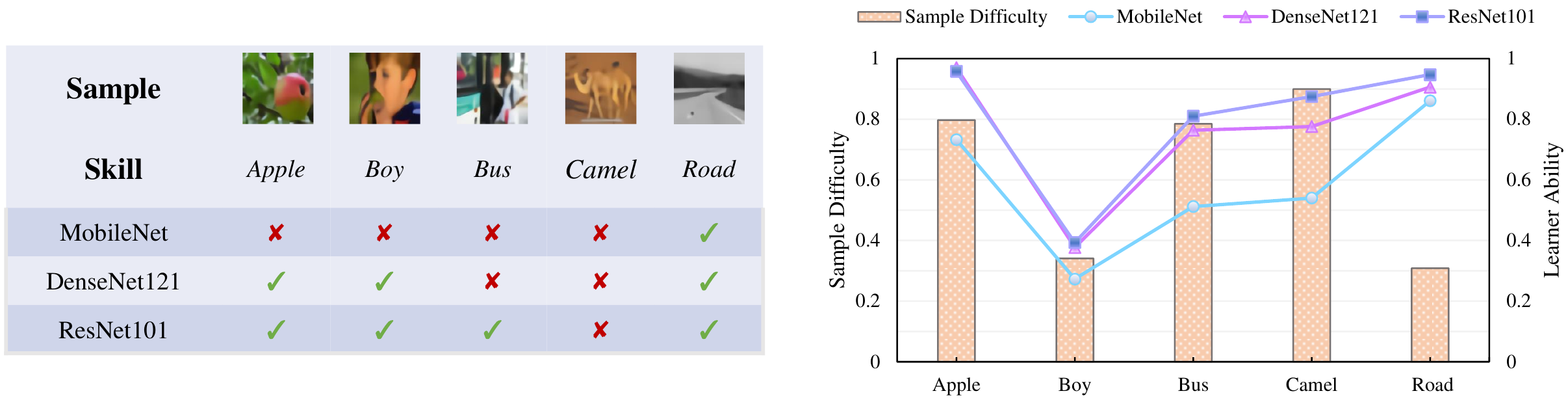}
  \caption{\textbf{Diagnosis examples of learners randomly selected from CIFAR-100.} The left is the response logs from 3 learners to 5 samples with different skills. The right shows the diagnosed samples' \emph{difficulty}~(bars) and the learners' \textit{Ability} on corresponding samples~(points)}

  \label{Fig. visualization3}
\end{figure*}

\begin{figure*}[t]
  \centering 
  \includegraphics[width=0.8\textwidth]{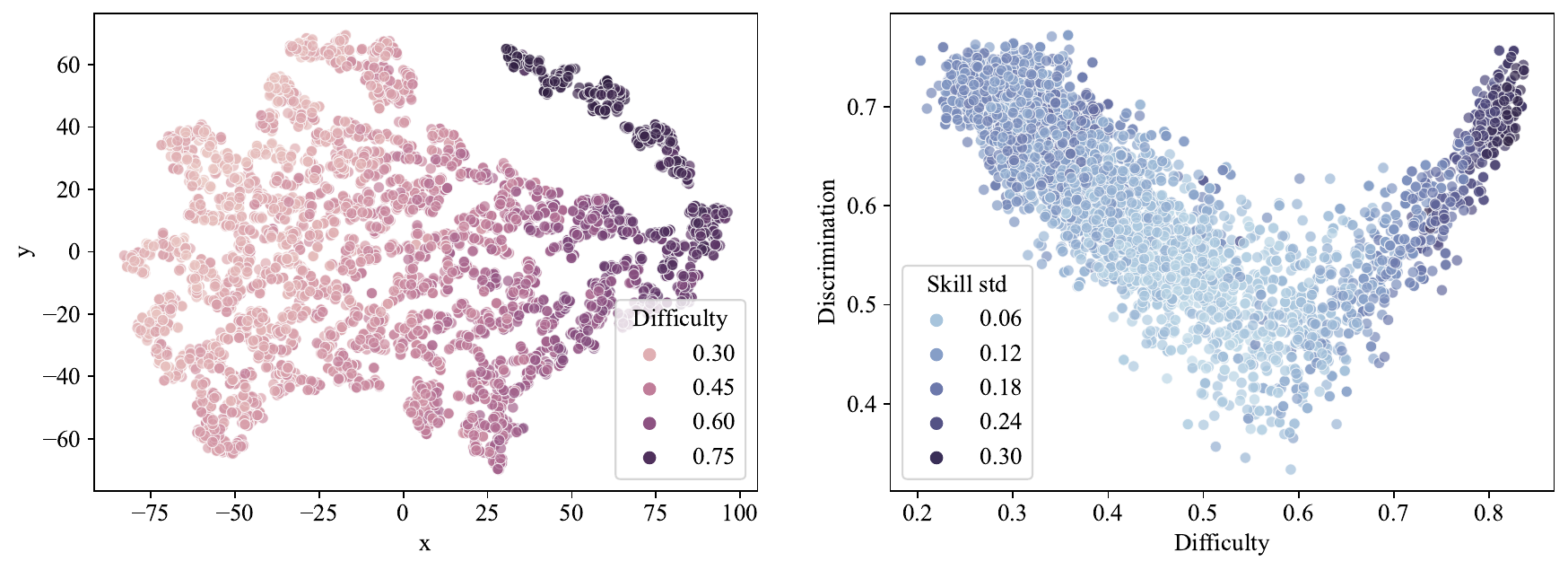}


  \caption{\textbf{Visualization of Camilla assessment cases in CIFAR-100.} The left part visualizes both the location of data samples in two dimensional \emph{skill mask} space and the color labels of sample \emph{difficulty}. The right part visualizes the sample \emph{difficulty}, \emph{discrimination} and the standard deviation of sample \emph{skill mask} factors.}

  

  \label{Fig.latent skill}
\end{figure*}

In this subsection, we intuitively visualize the factors~(i.e. learner \emph{Ability} and sample \emph{difficulty}) diagnosed by our diagnoser \emph{Camilla-Base} in Figure~\ref{Fig. visualization1}, Figure~\ref{Fig. visualization2} and Figure~\ref{Fig. visualization3}. For ease of understanding, we rescale the learner \emph{Ability} and sample factors into the range of [0, 1] via Sigmoid function. Then, we show the relations among latent \emph{skill mask} factors, sample \emph{difficulty} and sample \emph{discrimination} factors diagnosed by \emph{Camilla} in Figure~\ref{Fig.latent skill}. 





\noindent\textbf{What is the difference between \emph{Ability} and coarse-grained metric \emph{Accuracy}?} As shown in the radar maps of Figure~\ref{Fig. visualization1}B, we compare \emph{Ability} and the average \emph{Accuracy} of ResNet101 in several image classes of CIFAR-100 and Random Forest in Titanic, respectively. Let's take ResNet101 as an example, we can see the values of \textit{Ability} and \emph{Accuracy} are quite different from each other on most of the classes/skills. In the left scatter diagram of Figure~\ref{Fig. visualization1}C, we further visualize the correlation of \emph{Ability} and \emph{Accuracy} of three learner backbones~(i.e. Random Forest~(RF), SGD Classifier~(SGD) and Decision Tree~(DT)) in Titanic, where each scatter represents one learner under a specific hyperparameter setting. Both \emph{Accuracy} and \emph{Ability} of SGD vary considerably among different hyperparameters, while Random Forest and its variants are more robust. Moreover, in the right scatter diagram of Figure~\ref{Fig. visualization1}C, we visualize the difference from a standard deviation perspective by calculating the standard deviations of both multidimensional learner \emph{Ability} $A_i$ and \emph{Accuracy} values of the learner on different sample classes respectively. The results intuitively demonstrate that the standard deviation of learner \emph{Ability} on each class varies greatly from \emph{Accuracy}. In the summary of Figure~\ref{Fig. visualization1}C, as our proposed framework can collaboratively measure the learners' abilities and sample difficulty, the results of \emph{Ability} metric is quite different from that of traditional \emph{coarse-grained} metric (i.e. \emph{Accuracy}).

\noindent\textbf{What is the relation between the \emph{difficulty} factor and the inherent features of sample?} In Figure~\ref{Fig. visualization2}A, we select several samples of ``bus'', ``elephant'' and ``sunflower'' classes from CIFAR-100, and the corresponding value of their \emph{difficulty} factors from our \textit{Camilla-Base}. Intuitively, the \emph{difficulty} factor is in direct proportion to the complexity of the inherent features of sample~(e.g. a complex scenario). For instance, the image of these three classes with lowest \emph{difficulty} values contain distinct object outlines, whereas the image of ``bus'' with highest \emph{difficulty} value may be due to the complex scenario.

\noindent\textbf{What is the intuitive difference of learners in terms of \emph{Ability}?} As shown in the radar maps of Figure~\ref{Fig. visualization2}B, we compare the different performance of learners on different skills. Without loss of generality, we choose ResNet101 and DensNet121 as the typical learners in CIFAR-100 and choose Random Forest and SGD Classifier as the typical learners in Titanic respectively, and we also use typical sample classes or features to represent skills. For instance, although DenseNet121 and ResNet101 have similar \emph{Accuracy} performance in the whole CIFAR-100 from the overall \emph{Accuracy} result in  Supplementary Table~\ref{Table. CIFAR100 learner}, DenseNet121 outperforms ResNet101 with a large margin in the skills including  ``baby", ``clock" and ``house", while performs worse in the ``lobster", ``shark" and ``plate". In the left scatter diagram of Figure~\ref{Fig. visualization2}C, we visualize the mean and standard deviation of multidimensional \emph{Ability} of different learners in Titanic, and we  discover that more powerful learners~(with higher \emph{Ability} value) have more dissimilar performance in each class~(the standard deviation of \emph{Ability} value is higher). In the right histogram of Figure~\ref{Fig. visualization2}C, we further show the \emph{difficulty} distributions of data samples correctly responded by SGD and Random Forest, where we can see that Random Forest correctly answers easy samples than SGD, while SGD gives more correct answers to difficult samples.

\noindent\textbf{What is the intuitive relation between learner \emph{Ability} and sample \emph{difficulty}?}
The \emph{Ability} of learners and the factors of samples~(e.g. \emph{difficulty}) could provide detailed information of responses from learners to samples. As shown in the left part of Figure~\ref{Fig. visualization3}, we randomly select the response logs of three learners to five samples with different skills. In the right part, the bars represent sample \emph{difficulty} and the points denote the learners' \emph{Ability}. We can observe from the figure that the learners are more likely to make correct responses to samples when its \textit{Ability} surpasses the \emph{difficulty} of samples. For instance, ResNet101 predicted the ``bus'' sample successfully and the \textit{Ability} of ResNet101 is higher than the \emph{difficulty} of the ``bus'' sample.

\noindent\textbf{What is the relation among sample \emph{difficulty}, \emph{discrimination} factors and the latent \emph{skill mask} factors diagnosed by \emph{Camilla}?} As shown in Figure~\ref{Fig.latent skill}, we randomly select 50 samples from each image class in CIFAR-100 and visualize their \emph{difficulty} factors, \emph{discrimination} factors and the latent \emph{skill mask} factors diagnosed by \emph{Camilla}. Specifically, in the left scatter diagram of Figure~\ref{Fig.latent skill}, we reduce the dimensions of sample \emph{skill mask} factors into two dimensions by t-SNE~\cite{t-SNE}, and then visualize both the location of each sample in this two-dimensional space and the color label of its \emph{difficulty}. We can see that the samples with similar \emph{difficulty} value/color are closer, indicating that the samples requiring similar skills have similar \emph{difficulty} values. Meanwhile, the right part of Figure~\ref{Fig.latent skill} visualizes the \emph{difficulty} and \emph{discrimination} factors of those samples, where we color each sample based on the standard deviation of its \emph{skill mask} factors. The results show that the higher the standard deviation of sample \emph{skill mask} factors, the higher \emph{difficulty} or \emph{discrimination} of the corresponding samples. Besides, the samples with medium \emph{difficulty} values tend to have the lowest \emph{discrimination} values. All these discoveries from case studies support that \emph{Camilla} can well capture the different characteristics of data samples with multiple factors.





\subsection*{Comprehensive Evaluation of the Quality of Ability}

In this subsection, we comprehensively evaluate the quality of our proposed diagnosers \emph{Camilla}s along with the novel diagnostic metric \emph{Ability} in three aspects: (1)~Since there is no ground-truth value of the learner \textit{Ability}, we indirectly evaluate the reliability of diagnosers based on their performance in predicting the response of learners on untried samples. (2)~We evaluate the consistency of coarse-grained metrics~(e.g. \emph{Accuracy} for classification tasks) with the rankings of learners output by diagnosers~(e.g. IRT and our proposed \emph{Camilla}s).  (3)~We evaluate the rank stability of diagnosers' metrics on mutually exclusive samples, i.e. two non-overlapping partitions of the same sample set.


\noindent\textbf{Comparison Methods.} We compare \emph{Camilla}s (both \emph{Camilla-base} and \emph{Camilla}) with a number of representative baselines including widely-used cognitive diagnosis models and intuitive methods which can be collectively referred as diagnosers.
For the first, we derive two statistical algorithms called \emph{Vanilla} (or \emph{Skill-Vanilla}) which consider the probability that the learners predict untried samples correctly is the same as the \emph{Accuracy} of learners on all tried samples (or a part of samples with same skills).
For the second, we compare \emph{Camilla}s with IRT~\cite{embretson2013item}, which is a widespread unidimensional cognitive diagnosis model~(CDM). As for multidimensional CDM, we select classic MIRT~\cite{reckase2009multidimensional} which characterizes the learner \emph{Ability} in latent space, and a probabilistic matrix factorization~(MF) method~\cite{mnih2008probabilistic} which projects both users and items into low-dimensional space by mining the user-item response logs. Finally, we compare \emph{Camilla}s with NeuralCD~\cite{wang2020neural}, a deep CDM that uses multiple layers to predict the probability that students answer exercises correctly. All these baselines are trained and tested in the same experimental settings with \emph{Camilla}s, and the details of both experimental setup and diagnosers' implementation are given in the section of Method.

\begin{table*}[t]
  \renewcommand\arraystretch{0.9}
  
  \centering
  \caption{Reliability evaluation of diagnosers in Titanic, CIFAR-100, ESOL and Diamond. }
  \setlength{\tabcolsep}{1.3mm}{
  \begin{tabular}{lcccccccccccc}
    \toprule
    \multirow{2}{*}{Methods} & \multicolumn{4}{c}{Titanic} & \multicolumn{4}{c}{CIFAR-100} & \multicolumn{2}{c}{ESOL} & \multicolumn{2}{c}{Diamond} \\
    \cmidrule(l{.6em}r{.0em}){2-5} \cmidrule(l{.6em}r{.0em}){6-9} \cmidrule(l{.6em}r{.0em}){10-11} \cmidrule(l{.6em}r{.0em}){12-13}
                             &\ \ \ $\uparrow$  ACC      &\ \ \  $\uparrow$  F1                     &\ \ \ $\uparrow$  AUC            &\ \ \     $\downarrow$  RMSE       &\ \ \  $\uparrow$  ACC       &\ \ \  $\uparrow$  F1           &\ \ \ $\uparrow$   AUC            &\ \ \  $\downarrow$  RMSE         &\ \ \  $\downarrow$  MAE &\ \ \  $\downarrow$  RMSE  &\ \ \  $\downarrow$  MAE &\ \ \  $\downarrow$  RMSE    \\
    \midrule
    Vanilla  &\ \ \   0.625  &\ \ \  0.507  &\ \ \ 0.542 &\ \ \ 0.433   &\ \ \   0.573         &\ \ \   0.531          &\ \ \   0.615          &\ \ \   0.462   &\ \ \ - & \ \ \ - &\ \ \ - &\ \ \ -     \\
    Skill-Vanilla  &\ \ \   0.668  &\ \ \  0.562  &\ \ \ 0.683 &\ \ \ 0.420   &\ \ \   0.602         &\ \ \   0.563          &\ \ \   0.699          &\ \ \   0.447    &\ \ \ - &\ \ \ - &\ \ \ - &\ \ \ -    \\
    IRT                    &\ \ \   0.855                          &\ \ \   0.781                    &\ \ \   0.914          &\ \ \   0.311          &\ \ \    0.828         &\ \ \   0.803          &\ \ \   0.880          &\ \ \   0.357   &\ \ \ 0.203 &\ \ \  0.264 &\ \ \ 0.152 &\ \ \ 0.209      \\
    MIRT                    &\ \ \   0.771                          &\ \ \   0.533                    &\ \ \   0.928          &\ \ \   0.345          &\ \ \   0.814          &\ \ \   0.772          &\ \ \   0.887          &\ \ \   0.375    &\ \ \ 0.201 &\ \ \ 0.260 &\ \ \ 0.105  &\ \ \ 0.152     \\
    MF                    &\ \ \   0.886                       &\ \ \   0.841                    &\ \ \   0.931   &\ \ \   0.292          &\ \ \   0.840          &\ \ \   0.821          &\ \ \   0.911          &\ \ \   0.336       &\ \ \  0.205  &  \ \ \ 0.261 & \ \ \ 0.132 & \ \ \ 0.183    \\
    NeuralCD                    &\ \ \   0.898                  &\ \ \  \textbf{0.869}                     &\ \ \   0.951          &\ \ \   0.270          &\ \ \   0.843          &\ \ \   0.825          &\ \ \   0.918         &\ \ \   0.331     & \ \ \ - & \ \ \ - & \ \ \ 0.108 & \ \ \ 0.154   \\
    \midrule
    Camilla-base  &\ \ \   0.893      &\ \ \   0.852     &\ \ \  0.948   &\ \ \    0.275          &\ \ \   0.832          &\ \ \   0.812         &\ \ \   0.896    &\ \ \   0.346     & \ \ \ - & \ \ \ - & \ \ \  0.092  & \ \ \  0.137  \\
    Camilla &\ \ \   \textbf{0.899}      &\ \ \   0.868  &\ \ \  \textbf{0.960}    &\ \ \   \textbf{0.267}   &\ \ \   \textbf{0.844}          &\ \ \   \textbf{0.829}          &\ \ \   \textbf{0.920}       &\ \ \   \textbf{0.330}      & \ \ \ \textbf{0.195} &\ \ \ \textbf{0.254} &\ \ \  \textbf{0.075}  &\ \ \  \textbf{0.114}  \\
    \bottomrule
  \end{tabular}}
  \label{Table. reliability}
\end{table*}

\begin{figure*}[t]
  \centering 
  \includegraphics[width=1.0\textwidth]{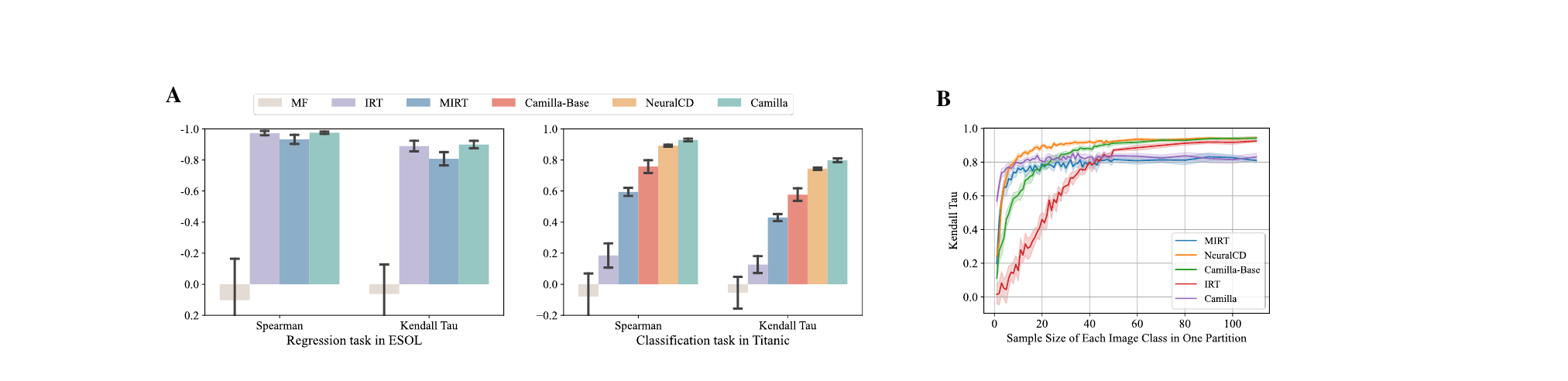}
  \caption{\textbf{A}, comparisons of the average$\pm$s.d. rank correlation between coarse-grained metrics~(\emph{MAE} for regression task in ESOL and \emph{Accuracy} for classification task in Titanic) and the assessment of diagnosers over ten independent runs in terms of \emph{Spearman correlation} and \emph{Kendall Tau}. The error bar indicates the standard deviation of rank correlation. \textbf{B}, comparison of the rank stability of diagnosers' metrics (MIRT, NeuralCD, \emph{Camilla-Base}, IRT and \emph{Camilla}) on mutually exclusive samples of CIFAR-100. Each number on the x-axis represents the sample size of each image class in one of the two partitions of CIFAR-100. Bands show the 95\% confidence intervals of \emph{Kendall Tau} correlations over ten independent runs per sample size.}

  \label{Fig.COR}
\end{figure*}

\noindent\textbf{How is the reliability of the diagnosers?} As the explanation given in Definition~\ref{def:diagnoser}, the performance of diagnosers in predicting the responses of learners on unknown/untried samples~(the prediction can be formulated as a classification problem or a regression problem) can reflect the reliability of each diagnoser. Then, for evaluating the performance of each diagnoser, we select the widely-used \emph{Accuracy}, root mean square error~(\emph{RMSE}), area under the curve~(\emph{AUC}) and macro \emph{F1}-score as evaluation metrics of classification tasks and \emph{RMSE} and mean absolute error~(\emph{MAE}) for regression tasks. We repeat all experiments ten times and report the average results of classification tasks and regression tasks in Table~\ref{Table. reliability}. There are several observations from these coarse-grained evaluation results. First, the response behaviors of each machine learner are actually predictable, and our \emph{Camilla} achieves the best or comparative predication performance on all four datasets, demonstrating \emph{Camilla} can effectively capture the pros and cons of machine learners. Second, the performance of our \emph{Camilla} and NeuralCD is near to each other in Titanic and CIFAR-100, whereas \emph{Camilla} can generalize to datasets without explicit skills with the help of latent skill mask factors. Therefore, \emph{Camilla} performs much better than others on the datasets of ESOL and Diamond. Last but not least, among the diagnosers except \emph{Camilla} and NeuralCD which both capture the interactions between learners and samples with neural networks (more specifically, MLP), \emph{Camilla-Base} generally performs the best. This illustrates the effectiveness of traditional cognitive diagnosis solutions where the learner behaviors are predicted via the comparison between her \emph{Ability} and the \emph{difficulty} of the sample skills~(Eq.~(\ref{equ:Camilla-Base})).

\noindent\textbf{How is the consistency of diagnosers and coarse-grained metric in ranking learners?} We rank learners according to the summarized \textit{Ability} value (the way of computing this value is given in the section of Method) and calculate the rank correlation with the order of learners ranked by coarse-grained metrics~(i.e. \emph{MAE} for regression task in ESOL and \emph{Accuracy} for classification task in Titanic).  
We suppose there should exist positive correlations between the rank orders.
Therefore, we select the widely-used \emph{Kendall's Tau} and \emph{Spearman's correlation} as evaluation metrics. We repeat this experiment ten times and report average results in Figure~\ref{Fig.COR}A.
We can see that our \emph{Camilla} has the best rank correlation and lower standard deviations with the coarse-grained metric than the metrics of traditional CDMs~(e.g. IRT, MIRT and NeuralCD), which also suggests reasonableness of both \emph{Camilla} and the \emph{Ability} metric. Note that, \emph{MAE} is negatively correlated with \emph{Ability}, and the lower rank correlation between \emph{MAE} and \emph{Ability} is the better. Besides, since MF characterizes samples and learners without the monotonicity assumption~(Assumption \ref{Ass:Mono}), the correlations between the assessment results of MF and coarse-grained metrics are explicitly lower than other CDMs, which illustrates that the monotonicity assumption ensures the positive correlation between the diagnosers' metrics and the proficiencies of learners.

\noindent\textbf{How is the rank stability of the diagnosers' metrics on mutually exclusive samples of the same dataset?} For a stable diagnostic metric (e.g. MIRT, NeuralCD, \emph{Camilla-Base}, IRT and \emph{Camilla}), the orders of different learners ranked by this metric should be consistent on different samples of the same dataset. For evaluating this rank stability, we randomly split all the data samples of one dataset (e.g. CIFAR-100) into two partitions (e.g. partition $A$ and partition $B$), and rank the learners on each partition based on the same diagnoser (e.g. rank the learners of ResNet, VGG and DenseNet on partitions $A$ and $B$ based on IRT, respectively). Then, we calculate the \emph{Kendall Tau} correlation between the two ranking lists of the learners on these two partitions. Therefore, the higher correlation the more stable diagnosing metric. The experimental results are visualized in Figure~\ref{Fig.COR}B, where we can have the following observations.
First, with the increasing number of samples in one partition, the learners' ranking correlations of all diagnosers become much higher. Second, the multi-dimensional CDMs have more reliable rankings than IRT in the cold start status~(i.e. the number of samples in each class is less than 40), demonstrating that the multidimensional \emph{Ability} are more stable in a small number of samples. Last but not least, \emph{Camilla-Base} and NeuralCD, which model explicit skills of the data sample, outperform other multi-dimensional CDMs in the convergence stage~(i.e. the number of samples in each class is larger than 80), illustrating the high quality of the explicit skills/classes labeled in CIFAR-100 data.

\section*{Related Work}
To the best of our knowledge, mainstream metrics generally evaluate machine learning algorithms from a rather simplistic unidimensional view, i.e. through summarizing learners' performance on single data samples. Without loss of generality, in the following, we take machine learning tasks of classification as an example for illustration. Specifically, for some typical and class-balanced classification datasets, such as CIFAR~\cite{cifar} and Imagenet~\cite{imagenet}, machine learning systems are usually ranked with Top-k \emph{Accuracy} classification score. While for the class-unbalanced classification tasks, mean average precision~(\emph{MAP}) and balanced \emph{F-score} are the widely used metrics. 

Considering the limitations of traditional coarse-grained metrics~\cite{japkowicz2006question, flach2019performance}, some recent \textit{interpretable evaluation} research~\cite{fu2020interpretable, moraffah2020causal} aim at dividing the holistic performance of the algorithms into interpretable groups of samples, which is denoted as \emph{post hoc dataset-level machine learning interpretation}~\cite{Murdoch2019interpretability}. For instance, ExplainaBoard~\cite{liu2021explainaboard} annotates a large number of attributes of samples in different natural language processing tasks and fine-grained analyses the performance of algorithms in different attributes. Similarly, Language Interpretability Tool~(LIT)~\cite{tenney2020language} provides visual explanations like attention weight and permits users real-time interact with machine learning algorithms to enable rapid error analysis. Besides, DIGEN~\cite{orzechowski2022generative} maximizes and evaluates the performance diversity of classification algorithms by designing a heuristic algorithm with mathematical functions and creating synthetic datasets with binary targets. In addition, inspired by the human Theory of Mind, some approaches~\cite{rabinowitz2018machine, cuzzolin2020knowing} seek to build a system to model agents' characteristics and mental states by predicting their future behavior, particularly in \emph{Reinforcement Learning}~(RL) field.

In contrast, cognitive diagnosis in psychometric measurement is developed to assess users' (e.g. students') proficiency on a set of tasks (e.g. exercises) or multiple skills through modeling their interactive behaviors~(e.g. historical exercise records), which can be viewed as one type of \emph{post hoc prediction-level interpretation}~\cite{Murdoch2019interpretability}. According to different diagnostic goals, researchers designed the corresponding cognitive diagnosis models~(CDMs) with two components: (1)~the representations of latent cognitive states and (2)~the interactive function. Based on the way of modeling latent cognitive states, the mainstream CDMs can be further divided into two categories: unidimension-oriented and multidimension-oriented. One prototype of the first type of CDMs is item response theory~(IRT)~\cite{lord1952theory, embretson2013item}, which characterizes students and exercises by a unidimensional variable respectively and predicts the probability that a student answers an exercise correctly as follows:
\begin{equation}
    P(r_{ij}=1|a_i, k_j, d_j) = \frac{1}{1 + e^{-1.7 k_j(a_i - d_j)}},
    \label{equ:IRT}
\end{equation}
 where $a_i$ denotes the unidimensional \emph{Ability} (cognitive state) of student $i$, $k_j$ and $d_j$ represent the \emph{discrimination} and \emph{difficulty} of exercise $j$, respectively. 

 The other type of CDMs aims to diagnose the students' proficiency in multiple skills. For instance, MIRT~\cite{reckase2009multidimensional} extends the $a_i$ and $k_j$ in IRT to multi-dimensional cognitive states. Also, Deterministic Input, Noisy ``And" gate model~(DINA)~\cite{de2009dina} describes each student with a binary vector that denotes whether the student masters a set of skills along with a Q-matrix which illustrates each exercise and relevant skills. Considering the limitations of the manually designed interaction functions of traditional CDMs (e.g. Eq.(\ref{equ:IRT})), NeuralCD~\cite{wang2020neural} combines deep neural networks to automatically learn such function:
 \begin{equation}
          x_{ij}=Q_j \circ\left(a_i-d_j\right) \times {k_j}, ~~
          P(r_{ij}=1) = \phi_n(...\phi_1(x_{ij})),
 \end{equation}
  where $\circ$ is element-wise product and $Q_j$ denotes the skills related to exercise $j$. $\phi_n(...\phi_1)$ indicates multiple mapping layers. 

Inspired by the psychometric theories, some recent \emph{interpretable evaluation} methods~\cite{martinez2016making, martinez2019item} apply the unidimensional metric of IRT in evaluating the  machine learning algorithms on specific tasks: question answering~\cite{rodriguez2021evaluation}, machine translation~\cite{hopkins-may-2013-models, otani-etal-2016-irt}, textual entailment~\cite{DBLP:conf/emnlp/LalorWY16, DBLP:conf/emnlp/LalorWY19}, chatbot evaluation~\cite{sedoc-ungar-2020-item} and AI video game~\cite{martinez2018dual} of RL. 

One step further from the unidimensional metric of IRT, in this paper, we aim to propose a task-agnostic evaluation framework, which can diagnose both multidimensional and unidimensional \emph{Ability} of machine learning algorithms via collaboratively quantifying both the abilities of different algorithms and some of the sample factors (e.g. sample \emph{difficulty}) in a general way. However, existing CDMs cannot be directly applied due to technical and domain challenges. For instance, the sample-skill relevancy (i.e. Q-matrix in psychometrics) is not always explicitly available in the machine learning tasks, in other words, we usually need to model the implicit skills for data samples. 

Similarly, some pioneer work~\cite{hernandez2017evaluation, osband2019behaviour} also appeal to the transformation of machine learning evaluation from task-oriented to ability-oriented. For instance, in the \emph{Bsuite}~\cite{osband2019behaviour} for \emph{Reinforcement Learning} evaluation, the agent performance of different tasks are aggregated into several core capabilities~(e.g. credit assignment and exploration). In this paper, we intuitively show that our proposed framework \emph{Camilla} can simultaneously quantify both the abilities of each algorithm on multiple skills and some of the sample factors~(e.g. sample \textit{difficulty}). From a cognitive diagnosis perspective, this framework may shed new light on the ability-oriented evaluation of multiple tasks, e.g. assigning common skill spaces to all the data samples.

Moreover, in the practical scenarios (e.g. the public leaderboards), if traditional unidimensional evaluation metrics work together with our proposed multi-dimensional metric \emph{Ability}, not only the strengths and weaknesses of each learner can be diagnosed but also the quality of the data samples can be quantified. Meanwhile, we can even precisely predict the response behaviors of each learner on unknown data samples
(e.g. the correct/wrong classification of one classifier) with high interpretability. Therefore, more adaptive training/testing solutions for the single learner can be designed, and more efficient ensemble learnings for multiple learners can be also achieved in the future.
Though \emph{Camilla} and
\emph{Ability} offer more sufficient information than traditional evaluation solutions, we should note that \emph{Camilla} incorporates a deep neural network to model the interactions between samples and learners where additional hyperparameters may undermine the stability of assessment results. Besides, the \emph{Ability} of each learner is evaluated in a collaborative way, which means the more learners in the response matrix $R$ the more precise evaluation of these learners' \emph{Ability} can be obtained by \emph{Camilla}.




\section*{Conclusions and Future Work}
In this paper, we proposed the task-agnostic cognitive diagnostic framework \textit{Camilla} and a multi-dimensional metric \textit{Ability} for providing both interpretable and reliable assessment of machine learning algorithms. To the best of our knowledge, this is the first comprehensive attempt for measuring the multifaceted strength of each machine learning algorithm through exploring the connections between the researches on psychometric theories and machine learning evaluation. Extensive experiments demonstrated the broad applicability and quantifiable reliability of our proposed diagnostic framework. Based on the interesting discoveries of this paper, we plan to further evaluate more algorithms with larger data. Meanwhile, we will adopt our solutions to help the downstream applications like self-adaptive training of machine learners.

\section*{Acknowledgements}
We would like to thank State Key Laboratory of Cognitive Intelligence for providing the computational resources for this research. We also would like to thank reviewers for their helpful discussions and  comments on the manuscript. This research was supported by grants from the National Natural Science Foundation of China (Grant No.s 61922073 and U20A20229).

\bibliographystyle{ACM-Reference-Format}
\bibliography{sample-base}


\begin{thebibliography}{70}


\ifx \showCODEN    \undefined \def \showCODEN     #1{\unskip}     \fi
\ifx \showDOI      \undefined \def \showDOI       #1{#1}\fi
\ifx \showISBNx    \undefined \def \showISBNx     #1{\unskip}     \fi
\ifx \showISBNxiii \undefined \def \showISBNxiii  #1{\unskip}     \fi
\ifx \showISSN     \undefined \def \showISSN      #1{\unskip}     \fi
\ifx \showLCCN     \undefined \def \showLCCN      #1{\unskip}     \fi
\ifx \shownote     \undefined \def \shownote      #1{#1}          \fi
\ifx \showarticletitle \undefined \def \showarticletitle #1{#1}   \fi
\ifx \showURL      \undefined \def \showURL       {\relax}        \fi
\providecommand\bibfield[2]{#2}
\providecommand\bibinfo[2]{#2}
\providecommand\natexlab[1]{#1}
\providecommand\showeprint[2][]{arXiv:#2}

\bibitem[Becht et~al\mbox{.}(2021)]%
        {becht2021high}
\bibfield{author}{\bibinfo{person}{Etienne Becht}, \bibinfo{person}{Daniel
  Tolstrup}, \bibinfo{person}{Charles-Antoine Dutertre},
  \bibinfo{person}{Peter~A Morawski}, \bibinfo{person}{Daniel~J Campbell},
  \bibinfo{person}{Florent Ginhoux}, \bibinfo{person}{Evan~W Newell},
  \bibinfo{person}{Raphael Gottardo}, {and} \bibinfo{person}{Mark~B Headley}.}
  \bibinfo{year}{2021}\natexlab{}.
\newblock \showarticletitle{High-throughput single-cell quantification of
  hundreds of proteins using conventional flow cytometry and machine learning}.
\newblock \bibinfo{journal}{\emph{Science Advances}} \bibinfo{volume}{7},
  \bibinfo{number}{39} (\bibinfo{year}{2021}), \bibinfo{pages}{eabg0505}.
\newblock


\bibitem[Bengio et~al\mbox{.}(2009)]%
        {bengio2009curriculum}
\bibfield{author}{\bibinfo{person}{Yoshua Bengio},
  \bibinfo{person}{J{\'e}r{\^o}me Louradour}, \bibinfo{person}{Ronan
  Collobert}, {and} \bibinfo{person}{Jason Weston}.}
  \bibinfo{year}{2009}\natexlab{}.
\newblock \showarticletitle{Curriculum learning}. In
  \bibinfo{booktitle}{\emph{Proceedings of the 26th annual international
  conference on machine learning}}. \bibinfo{pages}{41--48}.
\newblock


\bibitem[Braunstein et~al\mbox{.}(2021)]%
        {braunstein2021expectation}
\bibfield{author}{\bibinfo{person}{Alfredo Braunstein}, \bibinfo{person}{Thomas
  Gueudr{\'e}}, \bibinfo{person}{Andrea Pagnani}, {and} \bibinfo{person}{Mirko
  Pieropan}.} \bibinfo{year}{2021}\natexlab{}.
\newblock \showarticletitle{Expectation propagation on the diluted Bayesian
  classifier}.
\newblock \bibinfo{journal}{\emph{Physical Review E}} \bibinfo{volume}{103},
  \bibinfo{number}{4} (\bibinfo{year}{2021}), \bibinfo{pages}{043301}.
\newblock


\bibitem[Browne(2000)]%
        {browne2000cross}
\bibfield{author}{\bibinfo{person}{Michael~W Browne}.}
  \bibinfo{year}{2000}\natexlab{}.
\newblock \showarticletitle{Cross-validation methods}.
\newblock \bibinfo{journal}{\emph{Journal of mathematical psychology}}
  \bibinfo{volume}{44}, \bibinfo{number}{1} (\bibinfo{year}{2000}),
  \bibinfo{pages}{108--132}.
\newblock


\bibitem[Cortes and Vapnik(1995)]%
        {cortes1995support}
\bibfield{author}{\bibinfo{person}{Corinna Cortes} {and}
  \bibinfo{person}{Vladimir Vapnik}.} \bibinfo{year}{1995}\natexlab{}.
\newblock \showarticletitle{Support-vector networks}.
\newblock \bibinfo{journal}{\emph{Machine learning}} \bibinfo{volume}{20},
  \bibinfo{number}{3} (\bibinfo{year}{1995}), \bibinfo{pages}{273--297}.
\newblock


\bibitem[Cuzzolin et~al\mbox{.}(2020)]%
        {cuzzolin2020knowing}
\bibfield{author}{\bibinfo{person}{Fabio Cuzzolin}, \bibinfo{person}{Andrea
  Morelli}, \bibinfo{person}{Bogdan Cirstea}, {and} \bibinfo{person}{Barbara~J
  Sahakian}.} \bibinfo{year}{2020}\natexlab{}.
\newblock \showarticletitle{Knowing me, knowing you: theory of mind in AI}.
\newblock \bibinfo{journal}{\emph{Psychological medicine}}
  \bibinfo{volume}{50}, \bibinfo{number}{7} (\bibinfo{year}{2020}),
  \bibinfo{pages}{1057--1061}.
\newblock


\bibitem[De~La~Torre(2009)]%
        {de2009dina}
\bibfield{author}{\bibinfo{person}{Jimmy De~La~Torre}.}
  \bibinfo{year}{2009}\natexlab{}.
\newblock \showarticletitle{DINA model and parameter estimation: A didactic}.
\newblock \bibinfo{journal}{\emph{Journal of educational and behavioral
  statistics}} \bibinfo{volume}{34}, \bibinfo{number}{1}
  (\bibinfo{year}{2009}), \bibinfo{pages}{115--130}.
\newblock


\bibitem[Delaney(2004)]%
        {delaney2004esol}
\bibfield{author}{\bibinfo{person}{John~S Delaney}.}
  \bibinfo{year}{2004}\natexlab{}.
\newblock \showarticletitle{ESOL: estimating aqueous solubility directly from
  molecular structure}.
\newblock \bibinfo{journal}{\emph{Journal of chemical information and computer
  sciences}} \bibinfo{volume}{44}, \bibinfo{number}{3} (\bibinfo{year}{2004}),
  \bibinfo{pages}{1000--1005}.
\newblock


\bibitem[Deng et~al\mbox{.}(2009)]%
        {imagenet}
\bibfield{author}{\bibinfo{person}{Jia Deng}, \bibinfo{person}{Wei Dong},
  \bibinfo{person}{Richard Socher}, \bibinfo{person}{Li-Jia Li},
  \bibinfo{person}{Kai Li}, {and} \bibinfo{person}{Li Fei-Fei}.}
  \bibinfo{year}{2009}\natexlab{}.
\newblock \showarticletitle{ImageNet: A large-scale hierarchical image
  database}. In \bibinfo{booktitle}{\emph{2009 IEEE Conference on Computer
  Vision and Pattern Recognition}}. \bibinfo{pages}{248--255}.
\newblock
\urldef\tempurl%
\url{https://doi.org/10.1109/CVPR.2009.5206848}
\showDOI{\tempurl}


\bibitem[DiBello et~al\mbox{.}(2006)]%
        {dibello200631a}
\bibfield{author}{\bibinfo{person}{Louis~V DiBello}, \bibinfo{person}{Louis~A
  Roussos}, {and} \bibinfo{person}{William Stout}.}
  \bibinfo{year}{2006}\natexlab{}.
\newblock \showarticletitle{31a review of cognitively diagnostic assessment and
  a summary of psychometric models}.
\newblock \bibinfo{journal}{\emph{Handbook of statistics}}
  \bibinfo{volume}{26} (\bibinfo{year}{2006}), \bibinfo{pages}{979--1030}.
\newblock


\bibitem[Drummond and Japkowicz(2010)]%
        {drummond2010warning}
\bibfield{author}{\bibinfo{person}{Chris Drummond} {and}
  \bibinfo{person}{Nathalie Japkowicz}.} \bibinfo{year}{2010}\natexlab{}.
\newblock \showarticletitle{Warning: statistical benchmarking is addictive.
  Kicking the habit in machine learning}.
\newblock \bibinfo{journal}{\emph{Journal of Experimental \& Theoretical
  Artificial Intelligence}} \bibinfo{volume}{22}, \bibinfo{number}{1}
  (\bibinfo{year}{2010}), \bibinfo{pages}{67--80}.
\newblock


\bibitem[Embretson and Reise(2013)]%
        {embretson2013item}
\bibfield{author}{\bibinfo{person}{Susan~E Embretson} {and}
  \bibinfo{person}{Steven~P Reise}.} \bibinfo{year}{2013}\natexlab{}.
\newblock \bibinfo{booktitle}{\emph{Item response theory}}.
\newblock \bibinfo{publisher}{Psychology Press}.
\newblock


\bibitem[Ethayarajh et~al\mbox{.}(2022)]%
        {ethayarajh2022understanding}
\bibfield{author}{\bibinfo{person}{Kawin Ethayarajh}, \bibinfo{person}{Yejin
  Choi}, {and} \bibinfo{person}{Swabha Swayamdipta}.}
  \bibinfo{year}{2022}\natexlab{}.
\newblock \showarticletitle{Understanding Dataset Difficulty with $\mathcal{V}
  $-Usable Information}. In \bibinfo{booktitle}{\emph{International Conference
  on Machine Learning}}. PMLR, \bibinfo{pages}{5988--6008}.
\newblock


\bibitem[Fix and Hodges(1989)]%
        {fix1989discriminatory}
\bibfield{author}{\bibinfo{person}{Evelyn Fix} {and}
  \bibinfo{person}{Joseph~Lawson Hodges}.} \bibinfo{year}{1989}\natexlab{}.
\newblock \showarticletitle{Discriminatory analysis. Nonparametric
  discrimination: Consistency properties}.
\newblock \bibinfo{journal}{\emph{International Statistical Review/Revue
  Internationale de Statistique}} \bibinfo{volume}{57}, \bibinfo{number}{3}
  (\bibinfo{year}{1989}), \bibinfo{pages}{238--247}.
\newblock


\bibitem[Flach(2019)]%
        {flach2019performance}
\bibfield{author}{\bibinfo{person}{Peter Flach}.}
  \bibinfo{year}{2019}\natexlab{}.
\newblock \showarticletitle{Performance evaluation in machine learning: the
  good, the bad, the ugly, and the way forward}. In
  \bibinfo{booktitle}{\emph{Proceedings of the AAAI Conference on Artificial
  Intelligence}}, Vol.~\bibinfo{volume}{33}. \bibinfo{pages}{9808--9814}.
\newblock


\bibitem[Fu et~al\mbox{.}(2020)]%
        {fu2020interpretable}
\bibfield{author}{\bibinfo{person}{Jinlan Fu}, \bibinfo{person}{Pengfei Liu},
  {and} \bibinfo{person}{Graham Neubig}.} \bibinfo{year}{2020}\natexlab{}.
\newblock \showarticletitle{Interpretable Multi-dataset Evaluation for Named
  Entity Recognition}. In \bibinfo{booktitle}{\emph{Proceedings of the 2020
  Conference on Empirical Methods in Natural Language Processing (EMNLP)}}.
  \bibinfo{pages}{6058--6069}.
\newblock


\bibitem[Geirhos et~al\mbox{.}(2020)]%
        {geirhos2020shortcut}
\bibfield{author}{\bibinfo{person}{Robert Geirhos},
  \bibinfo{person}{J{\"o}rn-Henrik Jacobsen}, \bibinfo{person}{Claudio
  Michaelis}, \bibinfo{person}{Richard Zemel}, \bibinfo{person}{Wieland
  Brendel}, \bibinfo{person}{Matthias Bethge}, {and} \bibinfo{person}{Felix~A
  Wichmann}.} \bibinfo{year}{2020}\natexlab{}.
\newblock \showarticletitle{Shortcut learning in deep neural networks}.
\newblock \bibinfo{journal}{\emph{Nature Machine Intelligence}}
  \bibinfo{volume}{2}, \bibinfo{number}{11} (\bibinfo{year}{2020}),
  \bibinfo{pages}{665--673}.
\newblock


\bibitem[Gosiewska et~al\mbox{.}(2022)]%
        {gosiewska2022interpretable}
\bibfield{author}{\bibinfo{person}{Alicja Gosiewska},
  \bibinfo{person}{Katarzyna Wo{\'z}nica}, {and} \bibinfo{person}{Przemys{\l}aw
  Biecek}.} \bibinfo{year}{2022}\natexlab{}.
\newblock \showarticletitle{Interpretable meta-score for model performance}.
\newblock \bibinfo{journal}{\emph{Nature Machine Intelligence}}
  \bibinfo{volume}{4}, \bibinfo{number}{9} (\bibinfo{year}{2022}),
  \bibinfo{pages}{792--800}.
\newblock


\bibitem[Graves et~al\mbox{.}(2017)]%
        {graves2017automated}
\bibfield{author}{\bibinfo{person}{Alex Graves}, \bibinfo{person}{Marc~G
  Bellemare}, \bibinfo{person}{Jacob Menick}, \bibinfo{person}{R{\'e}mi Munos},
  {and} \bibinfo{person}{Koray Kavukcuoglu}.} \bibinfo{year}{2017}\natexlab{}.
\newblock \showarticletitle{Automated curriculum learning for neural networks}.
  In \bibinfo{booktitle}{\emph{Proceedings of the 34th International Conference
  on Machine Learning-Volume 70}}. \bibinfo{pages}{1311--1320}.
\newblock


\bibitem[He et~al\mbox{.}(2016)]%
        {he2016deep}
\bibfield{author}{\bibinfo{person}{Kaiming He}, \bibinfo{person}{Xiangyu
  Zhang}, \bibinfo{person}{Shaoqing Ren}, {and} \bibinfo{person}{Jian Sun}.}
  \bibinfo{year}{2016}\natexlab{}.
\newblock \showarticletitle{Deep residual learning for image recognition}. In
  \bibinfo{booktitle}{\emph{Proceedings of the IEEE conference on computer
  vision and pattern recognition}}. \bibinfo{pages}{770--778}.
\newblock


\bibitem[Hern{\'a}ndez-Orallo(2017)]%
        {hernandez2017evaluation}
\bibfield{author}{\bibinfo{person}{Jos{\'e} Hern{\'a}ndez-Orallo}.}
  \bibinfo{year}{2017}\natexlab{}.
\newblock \showarticletitle{Evaluation in artificial intelligence: from
  task-oriented to ability-oriented measurement}.
\newblock \bibinfo{journal}{\emph{Artificial Intelligence Review}}
  \bibinfo{volume}{48}, \bibinfo{number}{3} (\bibinfo{year}{2017}),
  \bibinfo{pages}{397--447}.
\newblock


\bibitem[Hern{\'a}ndez-Orallo et~al\mbox{.}(2021)]%
        {hernandez2021general}
\bibfield{author}{\bibinfo{person}{Jos{\'e} Hern{\'a}ndez-Orallo},
  \bibinfo{person}{Bao~Sheng Loe}, \bibinfo{person}{Lucy Cheke},
  \bibinfo{person}{Fernando Mart{\'\i}nez-Plumed}, {and}
  \bibinfo{person}{Se{\'a}n {\'O}~h{\'E}igeartaigh}.}
  \bibinfo{year}{2021}\natexlab{}.
\newblock \showarticletitle{General intelligence disentangled via a generality
  metric for natural and artificial intelligence}.
\newblock \bibinfo{journal}{\emph{Scientific reports}} \bibinfo{volume}{11},
  \bibinfo{number}{1} (\bibinfo{year}{2021}), \bibinfo{pages}{22822}.
\newblock


\bibitem[Hopkins and May(2013)]%
        {hopkins-may-2013-models}
\bibfield{author}{\bibinfo{person}{Mark Hopkins} {and}
  \bibinfo{person}{Jonathan May}.} \bibinfo{year}{2013}\natexlab{}.
\newblock \showarticletitle{Models of Translation Competitions}. In
  \bibinfo{booktitle}{\emph{Proceedings of the 51st Annual Meeting of the
  Association for Computational Linguistics (Volume 1: Long Papers)}}.
  \bibinfo{publisher}{Association for Computational Linguistics},
  \bibinfo{address}{Sofia, Bulgaria}, \bibinfo{pages}{1416--1424}.
\newblock


\bibitem[Hu et~al\mbox{.}(2020)]%
        {DBLP:conf/nips/HuFZDRLCL20}
\bibfield{author}{\bibinfo{person}{Weihua Hu}, \bibinfo{person}{Matthias Fey},
  \bibinfo{person}{Marinka Zitnik}, \bibinfo{person}{Yuxiao Dong},
  \bibinfo{person}{Hongyu Ren}, \bibinfo{person}{Bowen Liu},
  \bibinfo{person}{Michele Catasta}, {and} \bibinfo{person}{Jure Leskovec}.}
  \bibinfo{year}{2020}\natexlab{}.
\newblock \showarticletitle{Open Graph Benchmark: Datasets for Machine Learning
  on Graphs}. In \bibinfo{booktitle}{\emph{Advances in Neural Information
  Processing Systems 33: Annual Conference on Neural Information Processing
  Systems 2020, NeurIPS 2020, December 6-12, 2020, virtual}},
  \bibfield{editor}{\bibinfo{person}{Hugo Larochelle},
  \bibinfo{person}{Marc'Aurelio Ranzato}, \bibinfo{person}{Raia Hadsell},
  \bibinfo{person}{Maria{-}Florina Balcan}, {and} \bibinfo{person}{Hsuan{-}Tien
  Lin}} (Eds.).
\newblock


\bibitem[Huang et~al\mbox{.}(2017)]%
        {huang2017densely}
\bibfield{author}{\bibinfo{person}{Gao Huang}, \bibinfo{person}{Zhuang Liu},
  \bibinfo{person}{Laurens Van Der~Maaten}, {and} \bibinfo{person}{Kilian~Q
  Weinberger}.} \bibinfo{year}{2017}\natexlab{}.
\newblock \showarticletitle{Densely connected convolutional networks}. In
  \bibinfo{booktitle}{\emph{Proceedings of the IEEE conference on computer
  vision and pattern recognition}}. \bibinfo{pages}{4700--4708}.
\newblock


\bibitem[Jablonka et~al\mbox{.}(2023)]%
        {jablonka2023machine}
\bibfield{author}{\bibinfo{person}{Kevin~Maik Jablonka},
  \bibinfo{person}{Charithea Charalambous}, \bibinfo{person}{Eva
  Sanchez~Fernandez}, \bibinfo{person}{Georg Wiechers},
  \bibinfo{person}{Juliana Monteiro}, \bibinfo{person}{Peter Moser},
  \bibinfo{person}{Berend Smit}, {and} \bibinfo{person}{Susana Garcia}.}
  \bibinfo{year}{2023}\natexlab{}.
\newblock \showarticletitle{Machine learning for industrial processes:
  Forecasting amine emissions from a carbon capture plant}.
\newblock \bibinfo{journal}{\emph{Science Advances}} \bibinfo{volume}{9},
  \bibinfo{number}{1} (\bibinfo{year}{2023}), \bibinfo{pages}{eadc9576}.
\newblock


\bibitem[Japkowicz(2006)]%
        {japkowicz2006question}
\bibfield{author}{\bibinfo{person}{Nathalie Japkowicz}.}
  \bibinfo{year}{2006}\natexlab{}.
\newblock \showarticletitle{Why question machine learning evaluation methods}.
  In \bibinfo{booktitle}{\emph{AAAI workshop on evaluation methods for machine
  learning}}. \bibinfo{pages}{6--11}.
\newblock


\bibitem[Jiang et~al\mbox{.}(2014)]%
        {jiang2014self}
\bibfield{author}{\bibinfo{person}{Lu Jiang}, \bibinfo{person}{Deyu Meng},
  \bibinfo{person}{Shoou-I Yu}, \bibinfo{person}{Zhenzhong Lan},
  \bibinfo{person}{Shiguang Shan}, {and} \bibinfo{person}{Alexander
  Hauptmann}.} \bibinfo{year}{2014}\natexlab{}.
\newblock \showarticletitle{Self-paced learning with diversity}.
\newblock \bibinfo{journal}{\emph{Advances in Neural Information Processing
  Systems}}  \bibinfo{volume}{27} (\bibinfo{year}{2014}),
  \bibinfo{pages}{2078--2086}.
\newblock


\bibitem[Ke et~al\mbox{.}(2017)]%
        {ke2017lightgbm}
\bibfield{author}{\bibinfo{person}{Guolin Ke}, \bibinfo{person}{Qi Meng},
  \bibinfo{person}{Thomas Finley}, \bibinfo{person}{Taifeng Wang},
  \bibinfo{person}{Wei Chen}, \bibinfo{person}{Weidong Ma},
  \bibinfo{person}{Qiwei Ye}, {and} \bibinfo{person}{Tie-Yan Liu}.}
  \bibinfo{year}{2017}\natexlab{}.
\newblock \showarticletitle{Lightgbm: A highly efficient gradient boosting
  decision tree}.
\newblock \bibinfo{journal}{\emph{Advances in neural information processing
  systems}}  \bibinfo{volume}{30} (\bibinfo{year}{2017}),
  \bibinfo{pages}{3146--3154}.
\newblock


\bibitem[Kingma and Ba(2014)]%
        {kingma2014adam}
\bibfield{author}{\bibinfo{person}{Diederik~P Kingma} {and}
  \bibinfo{person}{Jimmy Ba}.} \bibinfo{year}{2014}\natexlab{}.
\newblock \showarticletitle{Adam: A method for stochastic optimization}.
\newblock \bibinfo{journal}{\emph{arXiv preprint arXiv:1412.6980}}
  (\bibinfo{year}{2014}).
\newblock


\bibitem[Kipf and Welling(2017)]%
        {DBLP:conf/iclr/KipfW17}
\bibfield{author}{\bibinfo{person}{Thomas~N. Kipf} {and} \bibinfo{person}{Max
  Welling}.} \bibinfo{year}{2017}\natexlab{}.
\newblock \showarticletitle{Semi-Supervised Classification with Graph
  Convolutional Networks}. In \bibinfo{booktitle}{\emph{5th International
  Conference on Learning Representations, {ICLR} 2017, Toulon, France, April
  24-26, 2017, Conference Track Proceedings}}.
\newblock


\bibitem[Krizhevsky et~al\mbox{.}(2009)]%
        {cifar}
\bibfield{author}{\bibinfo{person}{Alex Krizhevsky}, \bibinfo{person}{Geoffrey
  Hinton}, {et~al\mbox{.}}} \bibinfo{year}{2009}\natexlab{}.
\newblock \showarticletitle{Learning multiple layers of features from tiny
  images}.
\newblock  (\bibinfo{year}{2009}).
\newblock


\bibitem[Kumar et~al\mbox{.}(2010)]%
        {kumar2010self}
\bibfield{author}{\bibinfo{person}{M Kumar}, \bibinfo{person}{Benjamin Packer},
  {and} \bibinfo{person}{Daphne Koller}.} \bibinfo{year}{2010}\natexlab{}.
\newblock \showarticletitle{Self-paced learning for latent variable models}.
\newblock \bibinfo{journal}{\emph{Advances in neural information processing
  systems}}  \bibinfo{volume}{23} (\bibinfo{year}{2010}),
  \bibinfo{pages}{1189--1197}.
\newblock


\bibitem[Lalor et~al\mbox{.}(2016)]%
        {DBLP:conf/emnlp/LalorWY16}
\bibfield{author}{\bibinfo{person}{John~P. Lalor}, \bibinfo{person}{Hao Wu},
  {and} \bibinfo{person}{Hong Yu}.} \bibinfo{year}{2016}\natexlab{}.
\newblock \showarticletitle{Building an Evaluation Scale using Item Response
  Theory}. In \bibinfo{booktitle}{\emph{Proceedings of the 2016 Conference on
  Empirical Methods in Natural Language Processing, {EMNLP} 2016, Austin,
  Texas, USA, November 1-4, 2016}}, \bibfield{editor}{\bibinfo{person}{Jian
  Su}, \bibinfo{person}{Xavier Carreras}, {and} \bibinfo{person}{Kevin Duh}}
  (Eds.). \bibinfo{publisher}{The Association for Computational Linguistics},
  \bibinfo{pages}{648--657}.
\newblock
\urldef\tempurl%
\url{https://doi.org/10.18653/v1/d16-1062}
\showDOI{\tempurl}


\bibitem[Lalor et~al\mbox{.}(2019)]%
        {DBLP:conf/emnlp/LalorWY19}
\bibfield{author}{\bibinfo{person}{John~P. Lalor}, \bibinfo{person}{Hao Wu},
  {and} \bibinfo{person}{Hong Yu}.} \bibinfo{year}{2019}\natexlab{}.
\newblock \showarticletitle{Learning Latent Parameters without Human Response
  Patterns: Item Response Theory with Artificial Crowds}. In
  \bibinfo{booktitle}{\emph{Proceedings of the 2019 Conference on Empirical
  Methods in Natural Language Processing and the 9th International Joint
  Conference on Natural Language Processing, {EMNLP-IJCNLP} 2019, Hong Kong,
  China, November 3-7, 2019}}, \bibfield{editor}{\bibinfo{person}{Kentaro
  Inui}, \bibinfo{person}{Jing Jiang}, \bibinfo{person}{Vincent Ng}, {and}
  \bibinfo{person}{Xiaojun Wan}} (Eds.). \bibinfo{publisher}{Association for
  Computational Linguistics}, \bibinfo{pages}{4248--4258}.
\newblock
\urldef\tempurl%
\url{https://doi.org/10.18653/v1/D19-1434}
\showDOI{\tempurl}


\bibitem[LeCun et~al\mbox{.}(2015)]%
        {lecun2015deep}
\bibfield{author}{\bibinfo{person}{Yann LeCun}, \bibinfo{person}{Yoshua
  Bengio}, {and} \bibinfo{person}{Geoffrey Hinton}.}
  \bibinfo{year}{2015}\natexlab{}.
\newblock \showarticletitle{Deep learning}.
\newblock \bibinfo{journal}{\emph{nature}} \bibinfo{volume}{521},
  \bibinfo{number}{7553} (\bibinfo{year}{2015}), \bibinfo{pages}{436--444}.
\newblock


\bibitem[Leist et~al\mbox{.}(2022)]%
        {leist2022mapping}
\bibfield{author}{\bibinfo{person}{Anja~K Leist}, \bibinfo{person}{Matthias
  Klee}, \bibinfo{person}{Jung~Hyun Kim}, \bibinfo{person}{David~H Rehkopf},
  \bibinfo{person}{St{\'e}phane~PA Bordas}, \bibinfo{person}{Graciela
  Muniz-Terrera}, {and} \bibinfo{person}{Sara Wade}.}
  \bibinfo{year}{2022}\natexlab{}.
\newblock \showarticletitle{Mapping of machine learning approaches for
  description, prediction, and causal inference in the social and health
  sciences}.
\newblock \bibinfo{journal}{\emph{Science Advances}} \bibinfo{volume}{8},
  \bibinfo{number}{42} (\bibinfo{year}{2022}), \bibinfo{pages}{eabk1942}.
\newblock


\bibitem[Li et~al\mbox{.}(2021)]%
        {li2021dgl}
\bibfield{author}{\bibinfo{person}{Mufei Li}, \bibinfo{person}{Jinjing Zhou},
  \bibinfo{person}{Jiajing Hu}, \bibinfo{person}{Wenxuan Fan},
  \bibinfo{person}{Yangkang Zhang}, \bibinfo{person}{Yaxin Gu}, {and}
  \bibinfo{person}{George Karypis}.} \bibinfo{year}{2021}\natexlab{}.
\newblock \showarticletitle{Dgl-lifesci: An open-source toolkit for deep
  learning on graphs in life science}.
\newblock \bibinfo{journal}{\emph{ACS omega}} \bibinfo{volume}{6},
  \bibinfo{number}{41} (\bibinfo{year}{2021}), \bibinfo{pages}{27233--27238}.
\newblock


\bibitem[Liang et~al\mbox{.}(2020)]%
        {liang2020xglue}
\bibfield{author}{\bibinfo{person}{Yaobo Liang}, \bibinfo{person}{Nan Duan},
  \bibinfo{person}{Yeyun Gong}, \bibinfo{person}{Ning Wu},
  \bibinfo{person}{Fenfei Guo}, \bibinfo{person}{Weizhen Qi},
  \bibinfo{person}{Ming Gong}, \bibinfo{person}{Linjun Shou},
  \bibinfo{person}{Daxin Jiang}, \bibinfo{person}{Guihong Cao},
  \bibinfo{person}{Xiaodong Fan}, \bibinfo{person}{Ruofei Zhang},
  \bibinfo{person}{Rahul Agrawal}, \bibinfo{person}{Edward Cui},
  \bibinfo{person}{Sining Wei}, \bibinfo{person}{Taroon Bharti},
  \bibinfo{person}{Ying Qiao}, \bibinfo{person}{Jiun-Hung Chen},
  \bibinfo{person}{Winnie Wu}, \bibinfo{person}{Shuguang Liu},
  \bibinfo{person}{Fan Yang}, \bibinfo{person}{Daniel Campos},
  \bibinfo{person}{Rangan Majumder}, {and} \bibinfo{person}{Ming Zhou}.}
  \bibinfo{year}{2020}\natexlab{}.
\newblock \showarticletitle{{XGLUE}: A New Benchmark Datasetfor Cross-lingual
  Pre-training, Understanding and Generation}. In
  \bibinfo{booktitle}{\emph{Proceedings of the 2020 Conference on Empirical
  Methods in Natural Language Processing (EMNLP)}}.
  \bibinfo{publisher}{Association for Computational Linguistics},
  \bibinfo{address}{Online}, \bibinfo{pages}{6008--6018}.
\newblock
\urldef\tempurl%
\url{https://doi.org/10.18653/v1/2020.emnlp-main.484}
\showDOI{\tempurl}


\bibitem[Liu et~al\mbox{.}(2021)]%
        {liu2021explainaboard}
\bibfield{author}{\bibinfo{person}{Pengfei Liu}, \bibinfo{person}{Jinlan Fu},
  \bibinfo{person}{Yang Xiao}, \bibinfo{person}{Weizhe Yuan},
  \bibinfo{person}{Shuaichen Chang}, \bibinfo{person}{Junqi Dai},
  \bibinfo{person}{Yixin Liu}, \bibinfo{person}{Zihuiwen Ye}, {and}
  \bibinfo{person}{Graham Neubig}.} \bibinfo{year}{2021}\natexlab{}.
\newblock \showarticletitle{{E}xplaina{B}oard: An Explainable Leaderboard for
  {NLP}}. In \bibinfo{booktitle}{\emph{Proceedings of the 59th Annual Meeting
  of the Association for Computational Linguistics and the 11th International
  Joint Conference on Natural Language Processing: System Demonstrations}}.
  \bibinfo{publisher}{Association for Computational Linguistics},
  \bibinfo{pages}{280--289}.
\newblock
\urldef\tempurl%
\url{https://doi.org/10.18653/v1/2021.acl-demo.34}
\showDOI{\tempurl}


\bibitem[Lord(1952)]%
        {lord1952theory}
\bibfield{author}{\bibinfo{person}{Frederic Lord}.}
  \bibinfo{year}{1952}\natexlab{}.
\newblock \showarticletitle{A theory of test scores.}
\newblock \bibinfo{journal}{\emph{Psychometric monographs}}
  (\bibinfo{year}{1952}).
\newblock


\bibitem[Martinez-Plumed and Hernandez-Orallo(2018)]%
        {martinez2018dual}
\bibfield{author}{\bibinfo{person}{Fernando Martinez-Plumed} {and}
  \bibinfo{person}{Jose Hernandez-Orallo}.} \bibinfo{year}{2018}\natexlab{}.
\newblock \showarticletitle{Dual indicators to analyze ai benchmarks:
  Difficulty, discrimination, ability, and generality}.
\newblock \bibinfo{journal}{\emph{IEEE Transactions on Games}}
  \bibinfo{volume}{12}, \bibinfo{number}{2} (\bibinfo{year}{2018}),
  \bibinfo{pages}{121--131}.
\newblock


\bibitem[Mart{\'\i}nez-Plumed et~al\mbox{.}(2016)]%
        {martinez2016making}
\bibfield{author}{\bibinfo{person}{Fernando Mart{\'\i}nez-Plumed},
  \bibinfo{person}{Ricardo~BC Prud{\^e}ncio}, \bibinfo{person}{Adolfo
  Mart{\'\i}nez-Us{\'o}}, {and} \bibinfo{person}{Jos{\'e}
  Hern{\'a}ndez-Orallo}.} \bibinfo{year}{2016}\natexlab{}.
\newblock \showarticletitle{Making sense of item response theory in machine
  learning}.
\newblock In \bibinfo{booktitle}{\emph{ECAI 2016}}. \bibinfo{publisher}{IOS
  Press}, \bibinfo{pages}{1140--1148}.
\newblock


\bibitem[Mart{\'\i}nez-Plumed et~al\mbox{.}(2019)]%
        {martinez2019item}
\bibfield{author}{\bibinfo{person}{Fernando Mart{\'\i}nez-Plumed},
  \bibinfo{person}{Ricardo~BC Prud{\^e}ncio}, \bibinfo{person}{Adolfo
  Mart{\'\i}nez-Us{\'o}}, {and} \bibinfo{person}{Jos{\'e}
  Hern{\'a}ndez-Orallo}.} \bibinfo{year}{2019}\natexlab{}.
\newblock \showarticletitle{Item response theory in AI: Analysing machine
  learning classifiers at the instance level}.
\newblock \bibinfo{journal}{\emph{Artificial Intelligence}}
  \bibinfo{volume}{271} (\bibinfo{year}{2019}), \bibinfo{pages}{18--42}.
\newblock


\bibitem[Mnih and Salakhutdinov(2008)]%
        {mnih2008probabilistic}
\bibfield{author}{\bibinfo{person}{Andriy Mnih} {and} \bibinfo{person}{Russ~R
  Salakhutdinov}.} \bibinfo{year}{2008}\natexlab{}.
\newblock \showarticletitle{Probabilistic matrix factorization}. In
  \bibinfo{booktitle}{\emph{Advances in neural information processing
  systems}}. \bibinfo{pages}{1257--1264}.
\newblock


\bibitem[Monsalve-Bravo et~al\mbox{.}(2022)]%
        {monsalve2022analysis}
\bibfield{author}{\bibinfo{person}{Gloria~M Monsalve-Bravo},
  \bibinfo{person}{Brodie~AJ Lawson}, \bibinfo{person}{Christopher Drovandi},
  \bibinfo{person}{Kevin Burrage}, \bibinfo{person}{Kevin~S Brown},
  \bibinfo{person}{Christopher~M Baker}, \bibinfo{person}{Sarah~A Vollert},
  \bibinfo{person}{Kerrie Mengersen}, \bibinfo{person}{Eve McDonald-Madden},
  {and} \bibinfo{person}{Matthew~P Adams}.} \bibinfo{year}{2022}\natexlab{}.
\newblock \showarticletitle{Analysis of sloppiness in model simulations:
  Unveiling parameter uncertainty when mathematical models are fitted to data}.
\newblock \bibinfo{journal}{\emph{Science Advances}} \bibinfo{volume}{8},
  \bibinfo{number}{38} (\bibinfo{year}{2022}), \bibinfo{pages}{eabm5952}.
\newblock


\bibitem[Moraffah et~al\mbox{.}(2020)]%
        {moraffah2020causal}
\bibfield{author}{\bibinfo{person}{Raha Moraffah}, \bibinfo{person}{Mansooreh
  Karami}, \bibinfo{person}{Ruocheng Guo}, \bibinfo{person}{Adrienne Raglin},
  {and} \bibinfo{person}{Huan Liu}.} \bibinfo{year}{2020}\natexlab{}.
\newblock \showarticletitle{Causal interpretability for machine
  learning-problems, methods and evaluation}.
\newblock \bibinfo{journal}{\emph{ACM SIGKDD Explorations Newsletter}}
  \bibinfo{volume}{22}, \bibinfo{number}{1} (\bibinfo{year}{2020}),
  \bibinfo{pages}{18--33}.
\newblock


\bibitem[Murdoch et~al\mbox{.}(2019)]%
        {Murdoch2019interpretability}
\bibfield{author}{\bibinfo{person}{W.~James Murdoch}, \bibinfo{person}{Chandan
  Singh}, \bibinfo{person}{Karl Kumbier}, \bibinfo{person}{Reza Abbasi-Asl},
  {and} \bibinfo{person}{Bin Yu}.} \bibinfo{year}{2019}\natexlab{}.
\newblock \showarticletitle{Definitions, methods, and applications in
  interpretable machine learning}.
\newblock \bibinfo{journal}{\emph{Proceedings of the National Academy of
  Sciences}} \bibinfo{volume}{116}, \bibinfo{number}{44}
  (\bibinfo{year}{2019}), \bibinfo{pages}{22071--22080}.
\newblock
\urldef\tempurl%
\url{https://doi.org/10.1073/pnas.1900654116}
\showDOI{\tempurl}
\showeprint{https://www.pnas.org/doi/pdf/10.1073/pnas.1900654116}


\bibitem[Newell et~al\mbox{.}(1972)]%
        {newell1972human}
\bibfield{author}{\bibinfo{person}{Allen Newell},
  \bibinfo{person}{Herbert~Alexander Simon}, {et~al\mbox{.}}}
  \bibinfo{year}{1972}\natexlab{}.
\newblock \bibinfo{booktitle}{\emph{Human problem solving}}.
  Vol.~\bibinfo{volume}{104}.
\newblock \bibinfo{publisher}{Prentice-hall Englewood Cliffs, NJ}.
\newblock


\bibitem[Nichols et~al\mbox{.}(2012)]%
        {nichols2012cognitively}
\bibfield{author}{\bibinfo{person}{Paul~D Nichols}, \bibinfo{person}{Susan~F
  Chipman}, {and} \bibinfo{person}{Robert~L Brennan}.}
  \bibinfo{year}{2012}\natexlab{}.
\newblock \bibinfo{booktitle}{\emph{Cognitively diagnostic assessment}}.
\newblock \bibinfo{publisher}{Routledge}.
\newblock


\bibitem[Orzechowski and Moore(2022)]%
        {orzechowski2022generative}
\bibfield{author}{\bibinfo{person}{Patryk Orzechowski} {and}
  \bibinfo{person}{Jason~H Moore}.} \bibinfo{year}{2022}\natexlab{}.
\newblock \showarticletitle{Generative and reproducible benchmarks for
  comprehensive evaluation of machine learning classifiers}.
\newblock \bibinfo{journal}{\emph{Science Advances}} \bibinfo{volume}{8},
  \bibinfo{number}{47} (\bibinfo{year}{2022}), \bibinfo{pages}{eabl4747}.
\newblock


\bibitem[Osband et~al\mbox{.}(2019)]%
        {osband2019behaviour}
\bibfield{author}{\bibinfo{person}{Ian Osband}, \bibinfo{person}{Yotam Doron},
  \bibinfo{person}{Matteo Hessel}, \bibinfo{person}{John Aslanides},
  \bibinfo{person}{Eren Sezener}, \bibinfo{person}{Andre Saraiva},
  \bibinfo{person}{Katrina McKinney}, \bibinfo{person}{Tor Lattimore},
  \bibinfo{person}{Csaba Szepesvari}, \bibinfo{person}{Satinder Singh},
  {et~al\mbox{.}}} \bibinfo{year}{2019}\natexlab{}.
\newblock \showarticletitle{Behaviour Suite for Reinforcement Learning}. In
  \bibinfo{booktitle}{\emph{International Conference on Learning
  Representations}}.
\newblock


\bibitem[Otani et~al\mbox{.}(2016)]%
        {otani-etal-2016-irt}
\bibfield{author}{\bibinfo{person}{Naoki Otani}, \bibinfo{person}{Toshiaki
  Nakazawa}, \bibinfo{person}{Daisuke Kawahara}, {and} \bibinfo{person}{Sadao
  Kurohashi}.} \bibinfo{year}{2016}\natexlab{}.
\newblock \showarticletitle{{IRT}-based Aggregation Model of Crowdsourced
  Pairwise Comparison for Evaluating Machine Translations}. In
  \bibinfo{booktitle}{\emph{Proceedings of the 2016 Conference on Empirical
  Methods in Natural Language Processing}}. \bibinfo{publisher}{Association for
  Computational Linguistics}, \bibinfo{address}{Austin, Texas},
  \bibinfo{pages}{511--520}.
\newblock
\urldef\tempurl%
\url{https://doi.org/10.18653/v1/D16-1049}
\showDOI{\tempurl}


\bibitem[Paszke et~al\mbox{.}(2019)]%
        {paszke2019pytorch}
\bibfield{author}{\bibinfo{person}{Adam Paszke}, \bibinfo{person}{Sam Gross},
  \bibinfo{person}{Francisco Massa}, \bibinfo{person}{Adam Lerer},
  \bibinfo{person}{James Bradbury}, \bibinfo{person}{Gregory Chanan},
  \bibinfo{person}{Trevor Killeen}, \bibinfo{person}{Zeming Lin},
  \bibinfo{person}{Natalia Gimelshein}, \bibinfo{person}{Luca Antiga},
  {et~al\mbox{.}}} \bibinfo{year}{2019}\natexlab{}.
\newblock \showarticletitle{Pytorch: An imperative style, high-performance deep
  learning library}.
\newblock \bibinfo{journal}{\emph{Advances in neural information processing
  systems}}  \bibinfo{volume}{32} (\bibinfo{year}{2019}),
  \bibinfo{pages}{8026--8037}.
\newblock


\bibitem[Quinlan(1987)]%
        {quinlan1987simplifying}
\bibfield{author}{\bibinfo{person}{J.~Ross Quinlan}.}
  \bibinfo{year}{1987}\natexlab{}.
\newblock \showarticletitle{Simplifying decision trees}.
\newblock \bibinfo{journal}{\emph{International journal of man-machine
  studies}} \bibinfo{volume}{27}, \bibinfo{number}{3} (\bibinfo{year}{1987}),
  \bibinfo{pages}{221--234}.
\newblock


\bibitem[Rabinowitz et~al\mbox{.}(2018)]%
        {rabinowitz2018machine}
\bibfield{author}{\bibinfo{person}{Neil Rabinowitz}, \bibinfo{person}{Frank
  Perbet}, \bibinfo{person}{Francis Song}, \bibinfo{person}{Chiyuan Zhang},
  \bibinfo{person}{SM~Ali Eslami}, {and} \bibinfo{person}{Matthew Botvinick}.}
  \bibinfo{year}{2018}\natexlab{}.
\newblock \showarticletitle{Machine theory of mind}. In
  \bibinfo{booktitle}{\emph{International conference on machine learning}}.
  PMLR, \bibinfo{pages}{4218--4227}.
\newblock


\bibitem[Reckase(2009)]%
        {reckase2009multidimensional}
\bibfield{author}{\bibinfo{person}{Mark~D Reckase}.}
  \bibinfo{year}{2009}\natexlab{}.
\newblock \showarticletitle{Multidimensional item response theory models}.
\newblock In \bibinfo{booktitle}{\emph{Multidimensional item response theory}}.
  \bibinfo{publisher}{Springer}, \bibinfo{pages}{79--112}.
\newblock


\bibitem[Rodriguez et~al\mbox{.}(2021)]%
        {rodriguez2021evaluation}
\bibfield{author}{\bibinfo{person}{Pedro Rodriguez}, \bibinfo{person}{Joe
  Barrow}, \bibinfo{person}{Alexander~Miserlis Hoyle}, \bibinfo{person}{John~P
  Lalor}, \bibinfo{person}{Robin Jia}, {and} \bibinfo{person}{Jordan
  Boyd-Graber}.} \bibinfo{year}{2021}\natexlab{}.
\newblock \showarticletitle{Evaluation Examples Are Not Equally Informative:
  How Should That Change NLP Leaderboards?}. In
  \bibinfo{booktitle}{\emph{Proceedings of the 59th Annual Meeting of the
  Association for Computational Linguistics and the 11th International Joint
  Conference on Natural Language Processing (Volume 1: Long Papers)}}.
  \bibinfo{pages}{4486--4503}.
\newblock


\bibitem[Rosenbaum(1984)]%
        {rosenbaum1984testing}
\bibfield{author}{\bibinfo{person}{Paul~R Rosenbaum}.}
  \bibinfo{year}{1984}\natexlab{}.
\newblock \showarticletitle{Testing the conditional independence and
  monotonicity assumptions of item response theory}.
\newblock \bibinfo{journal}{\emph{Psychometrika}} \bibinfo{volume}{49},
  \bibinfo{number}{3} (\bibinfo{year}{1984}), \bibinfo{pages}{425--435}.
\newblock


\bibitem[Sedoc and Ungar(2020)]%
        {sedoc-ungar-2020-item}
\bibfield{author}{\bibinfo{person}{Jo{\~a}o Sedoc} {and} \bibinfo{person}{Lyle
  Ungar}.} \bibinfo{year}{2020}\natexlab{}.
\newblock \showarticletitle{Item Response Theory for Efficient Human Evaluation
  of Chatbots}. In \bibinfo{booktitle}{\emph{Proceedings of the First Workshop
  on Evaluation and Comparison of NLP Systems}}.
  \bibinfo{publisher}{Association for Computational Linguistics},
  \bibinfo{address}{Online}, \bibinfo{pages}{21--33}.
\newblock
\urldef\tempurl%
\url{https://doi.org/10.18653/v1/2020.eval4nlp-1.3}
\showDOI{\tempurl}


\bibitem[Sharifi-Noghabi et~al\mbox{.}(2021)]%
        {sharifi2021out}
\bibfield{author}{\bibinfo{person}{Hossein Sharifi-Noghabi},
  \bibinfo{person}{Parsa~Alamzadeh Harjandi}, \bibinfo{person}{Olga
  Zolotareva}, \bibinfo{person}{Colin~C Collins}, {and} \bibinfo{person}{Martin
  Ester}.} \bibinfo{year}{2021}\natexlab{}.
\newblock \showarticletitle{Out-of-distribution generalization from labelled
  and unlabelled gene expression data for drug response prediction}.
\newblock \bibinfo{journal}{\emph{Nature Machine Intelligence}}
  \bibinfo{volume}{3}, \bibinfo{number}{11} (\bibinfo{year}{2021}),
  \bibinfo{pages}{962--972}.
\newblock


\bibitem[Simonyan and Zisserman(2015)]%
        {DBLP:journals/corr/SimonyanZ14a}
\bibfield{author}{\bibinfo{person}{Karen Simonyan} {and}
  \bibinfo{person}{Andrew Zisserman}.} \bibinfo{year}{2015}\natexlab{}.
\newblock \showarticletitle{Very Deep Convolutional Networks for Large-Scale
  Image Recognition}. In \bibinfo{booktitle}{\emph{3rd International Conference
  on Learning Representations, {ICLR} 2015, San Diego, CA, USA, May 7-9, 2015,
  Conference Track Proceedings}}, \bibfield{editor}{\bibinfo{person}{Yoshua
  Bengio} {and} \bibinfo{person}{Yann LeCun}} (Eds.).
\newblock


\bibitem[Song et~al\mbox{.}(2021)]%
        {DBLP:conf/acl/00020ZZ020}
\bibfield{author}{\bibinfo{person}{Haoyu Song}, \bibinfo{person}{Yan Wang},
  \bibinfo{person}{Kaiyan Zhang}, \bibinfo{person}{Wei{-}Nan Zhang}, {and}
  \bibinfo{person}{Ting Liu}.} \bibinfo{year}{2021}\natexlab{}.
\newblock \showarticletitle{BoB: {BERT} Over {BERT} for Training Persona-based
  Dialogue Models from Limited Personalized Data}. In
  \bibinfo{booktitle}{\emph{Proceedings of the 59th Annual Meeting of the
  Association for Computational Linguistics and the 11th International Joint
  Conference on Natural Language Processing, {ACL/IJCNLP} 2021, (Volume 1: Long
  Papers), Virtual Event, August 1-6, 2021}},
  \bibfield{editor}{\bibinfo{person}{Chengqing Zong}, \bibinfo{person}{Fei
  Xia}, \bibinfo{person}{Wenjie Li}, {and} \bibinfo{person}{Roberto Navigli}}
  (Eds.). \bibinfo{publisher}{Association for Computational Linguistics},
  \bibinfo{pages}{167--177}.
\newblock
\urldef\tempurl%
\url{https://doi.org/10.18653/v1/2021.acl-long.14}
\showDOI{\tempurl}


\bibitem[Tatsuoka(1995)]%
        {tatsuoka1995architecture}
\bibfield{author}{\bibinfo{person}{Kikumi~K Tatsuoka}.}
  \bibinfo{year}{1995}\natexlab{}.
\newblock \showarticletitle{Architecture of knowledge structures and cognitive
  diagnosis: A statistical pattern recognition and classification approach}.
\newblock \bibinfo{journal}{\emph{Cognitively diagnostic assessment}}
  (\bibinfo{year}{1995}), \bibinfo{pages}{327--359}.
\newblock


\bibitem[Tenney et~al\mbox{.}(2020)]%
        {tenney2020language}
\bibfield{author}{\bibinfo{person}{Ian Tenney}, \bibinfo{person}{James Wexler},
  \bibinfo{person}{Jasmijn Bastings}, \bibinfo{person}{Tolga Bolukbasi},
  \bibinfo{person}{Andy Coenen}, \bibinfo{person}{Sebastian Gehrmann},
  \bibinfo{person}{Ellen Jiang}, \bibinfo{person}{Mahima Pushkarna},
  \bibinfo{person}{Carey Radebaugh}, \bibinfo{person}{Emily Reif},
  {et~al\mbox{.}}} \bibinfo{year}{2020}\natexlab{}.
\newblock \showarticletitle{The Language Interpretability Tool: Extensible,
  Interactive Visualizations and Analysis for NLP Models}. In
  \bibinfo{booktitle}{\emph{Proceedings of the 2020 Conference on Empirical
  Methods in Natural Language Processing: System Demonstrations}}.
  \bibinfo{pages}{107--118}.
\newblock


\bibitem[Udrescu and Tegmark(2020)]%
        {udrescu2020ai}
\bibfield{author}{\bibinfo{person}{Silviu-Marian Udrescu} {and}
  \bibinfo{person}{Max Tegmark}.} \bibinfo{year}{2020}\natexlab{}.
\newblock \showarticletitle{AI Feynman: A physics-inspired method for symbolic
  regression}.
\newblock \bibinfo{journal}{\emph{Science Advances}} \bibinfo{volume}{6},
  \bibinfo{number}{16} (\bibinfo{year}{2020}), \bibinfo{pages}{eaay2631}.
\newblock


\bibitem[van~der Maaten and Hinton(2008)]%
        {t-SNE}
\bibfield{author}{\bibinfo{person}{Laurens van~der Maaten} {and}
  \bibinfo{person}{Geoffrey Hinton}.} \bibinfo{year}{2008}\natexlab{}.
\newblock \showarticletitle{Visualizing Data using t-SNE}.
\newblock \bibinfo{journal}{\emph{Journal of Machine Learning Research}}
  \bibinfo{volume}{9}, \bibinfo{number}{86} (\bibinfo{year}{2008}),
  \bibinfo{pages}{2579--2605}.
\newblock


\bibitem[Veli{\v{c}}kovi{\'c} et~al\mbox{.}(2018)]%
        {velivckovic2018graph}
\bibfield{author}{\bibinfo{person}{Petar Veli{\v{c}}kovi{\'c}},
  \bibinfo{person}{Guillem Cucurull}, \bibinfo{person}{Arantxa Casanova},
  \bibinfo{person}{Adriana Romero}, \bibinfo{person}{Pietro Li{\`o}}, {and}
  \bibinfo{person}{Yoshua Bengio}.} \bibinfo{year}{2018}\natexlab{}.
\newblock \showarticletitle{Graph Attention Networks}. In
  \bibinfo{booktitle}{\emph{International Conference on Learning
  Representations}}.
\newblock


\bibitem[Voudouris et~al\mbox{.}(2022)]%
        {voudouris2022direct}
\bibfield{author}{\bibinfo{person}{Konstantinos Voudouris},
  \bibinfo{person}{Matthew Crosby}, \bibinfo{person}{Benjamin Beyret},
  \bibinfo{person}{Jos{\'e} Hern{\'a}ndez-Orallo}, \bibinfo{person}{Murray
  Shanahan}, \bibinfo{person}{Marta Halina}, {and} \bibinfo{person}{Lucy~G
  Cheke}.} \bibinfo{year}{2022}\natexlab{}.
\newblock \showarticletitle{Direct human-AI comparison in the animal-AI
  environment}.
\newblock \bibinfo{journal}{\emph{Frontiers in Psychology}}
  (\bibinfo{year}{2022}), \bibinfo{pages}{1884}.
\newblock


\bibitem[Wang et~al\mbox{.}(2020)]%
        {wang2020neural}
\bibfield{author}{\bibinfo{person}{Fei Wang}, \bibinfo{person}{Qi Liu},
  \bibinfo{person}{Enhong Chen}, \bibinfo{person}{Zhenya Huang},
  \bibinfo{person}{Yuying Chen}, \bibinfo{person}{Yu Yin}, \bibinfo{person}{Zai
  Huang}, {and} \bibinfo{person}{Wang Shijin}.}
  \bibinfo{year}{2020}\natexlab{}.
\newblock \showarticletitle{Neural cognitive diagnosis for intelligent
  education systems}. In \bibinfo{booktitle}{\emph{Proceedings of the AAAI
  Conference on Artificial Intelligence}}, Vol.~\bibinfo{volume}{34}.
  \bibinfo{pages}{6153--6161}.
\newblock


\end{thebibliography}


\clearpage

\section*{Data availability}
All data used in this paper are publicly available and can be accessed by clicking \href{https://www.kaggle.com/c/titanic}{here} for Titanic dataset, \href{https://www.cs.toronto.edu/~kriz/cifar.html}{here} for CIFAR-100 dataset, \href{https://lifesci.dgl.ai/api/data.html?highlight=dataset}{DGL-LifeSci}~\cite{li2021dgl} toolkit for ESOL dataset and \href{https://www.kaggle.com/shivam2503/diamonds}{here} for Diamond dataset. The response logs from our trained learners to each dataset samples can be assessed at \href{https://github.com/KellyGong/Camilla}{this Github repository}.

\section*{Code availability}
The source code can be publicly accessed at this \href{https://github.com/KellyGong/Camilla}{Github repository}. The training code of all learners in four datasets are released at \href{https://github.com/KellyGong/Camilla-Learner-Training}{https://github.com/KellyGong/Camilla-Learner-Training}. \newline

\section*{Implementation detail}
\noindent\textbf{Hyperparameters.}
We consider a wide range of values for each hyperparameter of the \emph{Camilla}s through random search on each dataset. Specifically, the value set of each hyperparameter is considered as follows: 

~~ Learning rate~(LR) = [0.0001, 0.001, 0.005, 0.01],

~~ First Layer dimension in MLP~(LD1) = [16, 32, 64, 128, 256, 512],

~~ First Layer dimension in MLP~(LD2) = [16, 32, 64],

~~ Latent skill dimension~(LS) = [5, 10, 20, 50].

For all datasets, we split the response logs into the training set, validation set and test set as 6:2:2. We use the validation set to optimize the values of these hyperparameters in the \emph{reliability evaluation} experiment~(Table~\ref{Table. reliability}) with mini-batch size 256. We select the optimal architecture of \emph{Camilla}s and visualize the results in the other experiments: \emph{diagnostic factors interpretation}~(Figure~\ref{Fig. visualization1}, Figure~\ref{Fig. visualization2} and Figure~\ref{Fig. visualization3}), \emph{rank consistency}~(Figure~\ref{Fig.COR}a) and \emph{rank stability}~(Figure~\ref{Fig.COR}b). Table~\ref{Table: statistics} summarizes the statistics related to these datasets, and for more information about the learners please refer to Supplementary.

\noindent\textbf{Architecture of Camilla.} The final hyperparameters and architectures of \emph{Camilla} for each dataset are as follows: 

~~\quad Dataset: \enspace\qquad\enspace LR, \quad LD1, \quad LD2, \quad LS

~~\quad Titanic: ~\qquad 0.001, ~\quad 128, ~~~\thinspace\quad 64, \quad\enspace 5

~~\quad CIFAR-100: ~~\thinspace 0.001, ~\quad 128, ~~~\thinspace\quad 64, \quad\enspace 5

~~\quad ESOL: ~~\thinspace\thinspace\qquad 0.001, ~\quad\thinspace\thinspace\thinspace 64, ~~~\thinspace\quad 32,  \quad 10

~~\quad Diamond: ~\thinspace\quad 0.001, ~\quad\thinspace\thinspace\thinspace 32, ~~~\thinspace\quad 16, \quad\enspace 5

LD1 and LD2 refer to the dimensions of the first layer and the second layer in the MLP respectively. For reproducing our experimental results, we independently run each experiment for ten times and consider the following random seeds:

~~\quad Seed = [1, 21, 42, 84, 168, 336, 672, 1344, 2688, 5376]

For the \emph{diagnostic factor interpretation} experiments, we choose 42 as the seed, which is the answer to life, the universe and everything~\cite{sharifi2021out}.

We optimize the performance of \emph{Camilla-Base} and other baselines with same settings and same ranges of values with \emph{Camilla} whenever those values are applicable. For the implementations of IRT, MIRT, NeuralCD and MF, we use the public code at \href{https://github.com/bigdata-ustc/EduCDM}{this url}. All of those diagnosers are implemented by PyTorch, and we adopt the Adam optimizer~\cite{kingma2014adam} to optimize the parameters of all those diagnosers whenever applicable.


\begin{table*}[b]
   \caption{Statistics of the experimental datasets.}
   \centering
   \renewcommand\arraystretch{1.05}
   \setlength{\tabcolsep}{2mm}{
   \begin{tabular}{lcccc}
   \toprule
   Statistics   & Titanic & CIFAR-100 & ESOL & Diamond \\ \midrule
   type   &  Classification  & Classification &  Regression &  Regression  \\
   explicit skill? & \cmark & \cmark & \xmark & \cmark \\
   \# samples   &  418  & 50,000 &  1,127 &  53,940  \\
   \# learners  &   353  & 42  &  22 & 157 \\
   \# response logs & 147,554 & 2,100,000 & 24,794 & 8,468,580 \\
   \bottomrule
   \end{tabular}
   \label{Table: statistics}
   }
\end{table*}


\noindent\textbf{Dataset Generation.} We generate the responses (i.e. response matrix $R$) of learners to the data samples of four datasets~(i.e. Titanic, CIFAR-100, ESOL and Diamond) by the following three steps. 

First of all, we implement popular and typical learners according to the characteristics of datasets. Let's take the image classification dataset~(i.e. CIFAR-100) as example. We reimplement 42 popular algorithms with top performance as learners, including ResNet~\cite{he2016deep}, VGG~\cite{DBLP:journals/corr/SimonyanZ14a} and DenseNet~\cite{huang2017densely} based on Pytorch~\cite{paszke2019pytorch}, and the list of learners for each dataset can be found in the \emph{Dataset Description} of section \emph{Results}. 

In the second step, we apply different strategies to train these learners on each dataset. As for CIFAR-100 and Diamond, we obtain the responses of all learners to each sample by 5-fold cross validation~\cite{browne2000cross}, i.e. train the learners on four-fifths samples and test the learners on the one-fifths samples, then iterate this process for five times until we obtain the test results on all samples as learners' responses. In ESOL, the learners with Graph Neural Network are pre-trained from Dgl-LifeSci~\cite{li2021dgl} toolkit and directly tested on the samples of ESOL. The responses from other learners in ESOL based on the fingerprints features of molecules to learners are obtained by 5-fold cross validation, which is same to CIFAR-100 and Diamond. Due to the small quantities of Titanic samples, we do not use cross validation which may lead to the unstable learner performance. Instead, we train the all learners in the external training set of Titanic, and test them on the 418 samples of Titanic. In summary, the characteristics~(e.g. hyperparameters and the size of learnable parameters) of these well-trained learners and their performance in Titanic, CIFAR-100, ESOL and Diamond are listed in Table~\ref{Table. titanic learner}, \ref{Table. CIFAR100 learner}, \ref{Table. ESOL learner} and \ref{Table. diamond learner}, respectively.

Finally, after getting the test results from well-trained learners to samples, the response matrix $R$ (input of our diagnoser) can be generated by selecting the learner $s_i$ and the sample $e_j$ as the indexes of $R$ and filling the value $r_{ij}$ derived from the test results with the transformation in the Section of \emph{Preliminary and Problem Definition}. For example, in the case of classification tasks, $r_{ij} = 1$ if learner $s_i$ answers the class label of sample $e_j$ correctly and $r_{ij} = 0$ otherwise.

\begin{table*}[h]
    \renewcommand\arraystretch{1.0}
    \centering
    \caption{Learner backbones, the values of hyperparameters and the statistics of learner performance in Titanic dataset.}
    \setlength{\tabcolsep}{1mm}{
    \begin{tabular}{lcccc}
      \hline
      \specialrule{0em}{2pt}{2pt}
      \textbf{Learner Backbone} & \textbf{Hyperparameters} & \textbf{\#~Learners} & \textbf{Acc mean} & \textbf{Acc std} \\
      \specialrule{0em}{2pt}{2pt}
      \hline
      \specialrule{0em}{2pt}{2pt}
      \multirow{4}{*}{Logistic Regression} & penalty: [l1, l2, elasticnet, none],  &   \multirow{4}{*}{4}   & \multirow{4}{*}{0.769} & \multirow{4}{*}{0.002} \\
         & C: [0.1, 0.5, 1, 5, 10],  &    &   &   \\
         & solver: [newton-cg, lbfgs, liblinear, sag, saga],  &   &   &  \\
         & warm start: [True, False]   &    &   &   \\
      \specialrule{0em}{2pt}{2pt}
      \hline
      \specialrule{0em}{2pt}{2pt}
      \multirow{3}{*}{Decision Tree} & criterion: [gini, entropy],  &   \multirow{3}{*}{12}   & \multirow{3}{*}{0.741} & \multirow{3}{*}{0.021} \\
         & splitter: [best, random],     &    &   &   \\
         & max features: [auto, sqrt, log2]  &   &   &  \\
      \specialrule{0em}{2pt}{2pt}
      \hline
      \specialrule{0em}{2pt}{2pt}
      \multirow{4}{*}{C-Support Vector} & kernel: [linear, poly, rbf, sigmoid, precomputed],  &   \multirow{4}{*}{15}   & \multirow{4}{*}{0.730} & \multirow{4}{*}{0.059} \\
         & gamma: [scale, auto],    &    &   &   \\
         & coef0: [0.0, 0.1, 1.0],  &   &   &  \\
         & tol: [0.01, 0.001, 0.0001]  &   &   &  \\
      \specialrule{0em}{2pt}{2pt}
      \hline
      \specialrule{0em}{2pt}{2pt}
      \multirow{4}{*}{Linear Support Vector} & penalty: [l1, l2],  &   \multirow{4}{*}{2}   & \multirow{4}{*}{0.766} & \multirow{4}{*}{0.0} \\
         & loss: [hinge, squared hinge],   &    &   &   \\
         & C: [.5, 2.0, 1.0],  &   &   &  \\
         & tol: [0.01, 0.001, 0.0001]  &   &   &  \\
      \specialrule{0em}{2pt}{2pt}
      \hline
      \specialrule{0em}{2pt}{2pt}
      \multirow{4}{*}{K-Nearest-Neighbors} & \#~neighbors: [3, 5, 7],  &   \multirow{4}{*}{20}   & \multirow{4}{*}{0.752} & \multirow{4}{*}{0.008} \\
         & weights: [uniform, distance],   &    &   &   \\
         & algorithm: [auto, ball tree, kd tree, brute],  &   &   &  \\
         & p: [1, 2]  &   &   &  \\
      \specialrule{0em}{2pt}{2pt}
      \hline
      \specialrule{0em}{2pt}{2pt}
      \multirow{5}{*}{Random Forest} & \#~estimators: [10, 50, 100],  &   \multirow{5}{*}{162}   & \multirow{5}{*}{0.779} & \multirow{5}{*}{0.007} \\
         & criterion: [gini, entropy],   &    &   &   \\
         & max depth: [5, 10, None],   &   &   &  \\
         & max features: [auto, sqrt, log2], &   &   &  \\
         & min samples leaf: [1, 3, 5]  &   &   &  \\
      \specialrule{0em}{2pt}{2pt}
      \hline
      \specialrule{0em}{2pt}{2pt}
      Gaussian Naive Bayes & var smoothing: [1e-9, 1e-8]  &   1   & 0.751 & 0.0 \\
      \specialrule{0em}{2pt}{2pt}
      \hline
      \specialrule{0em}{2pt}{2pt}
      \multirow{3}{*}{Perceptron} & penalty: [l2, l1, elasticnet],  &   \multirow{3}{*}{15}   & \multirow{3}{*}{0.689} & \multirow{3}{*}{0.04} \\
         & eta0: [1.0, 0.1, 0.5],   &    &   &   \\
         & early stopping: [True, False]   &   &   &  \\
      \specialrule{0em}{2pt}{2pt}
      \hline
      \specialrule{0em}{2pt}{2pt}
      \multirow{5}{*}{SGD Classifier} & loss: [hinge, log, modified huber,  &   \multirow{5}{*}{122}   & \multirow{5}{*}{0.710} & \multirow{5}{*}{0.066} \\
         & squared hinge, perceptron],   &    &   &   \\
         & penalty: [l1, l2, elasticnet],   &    &   &   \\
         & learning rate: [constant, optimal, invscaling, adaptive], &   &   &  \\
         & eta0: [0.1, 0.001, 0.0001] &   &   &  \\
      \specialrule{0em}{2pt}{2pt}
      \hline
      \specialrule{0em}{2pt}{2pt}
    \end{tabular}}
    \label{Table. titanic learner}
\end{table*}

\begin{table*}[h]
    \renewcommand\arraystretch{0.9}
    \centering
    \caption{Learners, the number of learnable parameters and performance statistics in CIFAR-100 dataset. }
    \setlength{\tabcolsep}{1.3mm}{
    \begin{tabular}{lcccc}
      \toprule
      \specialrule{0em}{1pt}{1pt}
      \textbf{Learner} & \textbf{\# learnable parameters~($\times 10^6$)} & \textbf{Acc mean} & \textbf{Acc std}   \\
      \specialrule{0em}{1pt}{1pt}
      \midrule
        ResNext101  & 25.3  &   0.780   &   0.095  \\ 
        DenseNet161  & 26.7  &   0.779   &   0.090  \\ 
        WideResnet  &  55.9 &   0.778   &   0.091  \\ 
        ResNext152  & 33.3  &   0.776   &   0.093  \\ 
        DenseNet201  & 18.3  &   0.775   &   0.088  \\ 
        ResNet152  & 58.3  &   0.773   &   0.091  \\ 
        ResNext50  & 14.8  &   0.770   &   0.091  \\ 
        DenseNet121  & 7.0  &   0.768   &   0.087  \\ 
        NasNet  &  5.2 &   0.766   &   0.090  \\ 
        Inception v3  &  22.3 &   0.765   &   0.085  \\ 
        ResNet101  & 42.7  &   0.762   &   0.091  \\ 
        Xception  & 21.0  &   0.761   &   0.088  \\ 
        ResNet50  & 23.7  &   0.759   &   0.087  \\ 
        SeResNet152  & 65.0  &   0.759   &   0.093  \\ 
        SeResNet101  & 47.5  &   0.758   &   0.091  \\ 
        ResNet34  & 21.3  &   0.757   &   0.088  \\ 
        StochasticDepth101  & 42.7  &   0.755   &   0.092  \\ 
        SeResNet34  & 21.5  &   0.755   &   0.090  \\ 
        SeResNet50  & 26.3  &   0.754   &   0.087  \\ 
        GoogleNet  & 6.4  &   0.752   &   0.085  \\ 
        SeResNet18  & 11.3  &   0.749   &   0.089  \\ 
        Inception v4  & 41.3  &   0.745   &   0.085  \\ 
        StochasticDepth34  & 21.3  &   0.745   &   0.085  \\ 
        ResNet18  & 11.2  &   0.743   &   0.086  \\ 
        PreactResnet101  & 42.7  &   0.742   &   0.087  \\ 
        PreactResnet152  &  58.4 &   0.742   &   0.087  \\ 
        PreactResnet50  & 23.7  &   0.738   &   0.085  \\ 
        StochasticDepth50  & 23.7  &   0.729   &   0.084  \\ 
        PreactResnet18  & 11.2  &   0.729   &   0.089  \\ 
        InceptionResnetv2  & 65.4  &   0.721   &   0.082  \\ 
        StochasticDepth18  & 11.2  &   0.708   &   0.084  \\ 
        VGG13  & 28.7  &   0.704   &   0.080  \\ 
        ShuffleNet  &  1.0 &   0.697   &   0.085  \\ 
        VGG16  &  34.0 &   0.696   &   0.079  \\ 
        ShuffleNetv2  &  1.4 &   0.695   &   0.084  \\ 
        SqueezeNet  &  0.8 &   0.682   &   0.085  \\ 
        MobileNetv2  & 2.4  &   0.672   &   0.076  \\ 
        VGG19  & 39.3  &   0.668   &   0.077  \\ 
        VGG11  & 28.5  &   0.664   &   0.079  \\ 
        MobileNet  & 3.3  &   0.657   &   0.082  \\ 
        Attention56  & 55.7  &   0.271   &   0.050  \\ 
        SVM  &  -  &   0.190   &   0.043  \\ 

      \bottomrule
    \end{tabular}}
    \label{Table. CIFAR100 learner}
\end{table*}

\begin{table*}[h]
   \renewcommand\arraystretch{1.2}
   \centering
   \caption{Learners and their performance statistics in ESOL dataset. The models without Graph Neural Network~(GNN) are trained on the fingerprints features of molecules, while the models with GNN are  trained on the graph structures of molecules.}
   \setlength{\tabcolsep}{1.3mm}{
   \begin{tabular}{lcccc}
     \toprule
     \specialrule{0em}{1pt}{1pt}
     \textbf{Learner} & \textbf{Graph Neural Network} & \textbf{MAE} & \textbf{RMSE}   \\
     \specialrule{0em}{1pt}{1pt}
     \midrule
      AttentiveFP\_canonical\_ESOL  & \cmark  &   0.425   &   0.580  \\ 
      MPNN\_canonical\_ESOL  & \cmark  &   0.483   &   0.655  \\ 
      MPNN\_attentivefp\_ESOL  & \cmark  &   0.497   &   0.668  \\ 
      AttentiveFP\_attentivefp\_ESOL  & \cmark  &   0.505   &   0.677  \\ 
      GIN\_supervised\_masking\_ESOL  & \cmark  &   0.517   &   0.753  \\ 
      Weave\_canonical\_ESOL  & \cmark  &   0.526   &   0.687  \\ 
      GIN\_supervised\_infomax\_ESOL  & \cmark  &   0.571   &   0.765  \\ 
      GCN\_canonical\_ESOL  & \cmark  &   0.581   &   0.770  \\ 
      GIN\_supervised\_contextpred\_ESOL  & \cmark  &   0.651   &   0.855  \\ 
      Weave\_attentivefp\_ESOL  & \cmark  &   0.716   &   0.951  \\ 
      CatBoost  & \xmark  &   0.887   &   1.196  \\ 
      GCN\_attentivefp\_ESOL  & \cmark  &   0.922   &   1.170  \\ 
      GAT\_attentivefp\_ESOL  & \cmark  &   0.936   &   1.174  \\ 
      Random Forest  & \xmark  &   0.959   &   1.295  \\ 
      GIN\_supervised\_edgepred\_ESOL  & \cmark  &   0.988   &   1.324  \\ 
      GradientBoosting  & \xmark  &   1.029   &   1.359  \\ 
      GAT\_canonical\_ESOL  & \cmark  &   1.053   &   1.306  \\ 
      K-Neighbors  & \xmark  &   1.156   &   1.547  \\ 
      Ridge  & \xmark  &   1.187   &   1.531  \\ 
      Linear Regression  & \xmark  &   1.216   &   1.565  \\ 
      ExtraTrees  & \xmark  &   1.280   &   1.785  \\ 
      AdaBoost  & \xmark  &   1.349   &   1.682  \\ 

     \bottomrule
   \end{tabular}}
   \label{Table. ESOL learner}
\end{table*}

\begin{table*}[h]
   \renewcommand\arraystretch{1.1}
   \centering
   \caption{Learner backbones, the values of hyperparameters and performance statistics in Diamond dataset. }
   \setlength{\tabcolsep}{1.3mm}{
   \begin{tabular}{lcccc}
     \toprule
     \specialrule{0em}{2pt}{2pt}
     \textbf{Learner Backbone} & \textbf{Hyperparameters} & \textbf{\#~Learners} & \textbf{MAE mean} & \textbf{MAE std}   \\
     \specialrule{0em}{2pt}{2pt}
     \midrule
     \specialrule{0em}{2pt}{2pt}
     Support Vector Regression & kernel: [rbf, linear]  &   2  &   413.8   &   70.5    \\
     \specialrule{0em}{2pt}{2pt}
     \midrule
     \specialrule{0em}{2pt}{2pt}
     Ridge & alpha=0.5, max\_iter=500, tol=0.001  &   1  &   743.9   &   0.0 \\
     \midrule
     \specialrule{0em}{2pt}{2pt}
     \multirow{2}{*}{Decision Tree Regressor} & min\_samples\_leaf: [2, 4, 8],   &   \multirow{2}{*}{9}   & \multirow{2}{*}{358.7} & \multirow{2}{*}{8.1} \\
        & min\_samples\_split: [2, 4, 8]     &    &   &   \\
      \specialrule{0em}{2pt}{2pt}
      \midrule
      \multirow{2}{*}{LightGBM} & max\_depth: [2, 4, 8],   &   \multirow{2}{*}{9}   & \multirow{2}{*}{346.8} & \multirow{2}{*}{70.2} \\
         & n\_estimators: [500, 1000, 2000]     &    &   &   \\
      \specialrule{0em}{2pt}{2pt}
      \midrule
      \specialrule{0em}{2pt}{2pt}
      \multirow{3}{*}{Random Forest} & max\_features: [2, 4, 8],   &   \multirow{3}{*}{18}   & \multirow{3}{*}{702.8} & \multirow{3}{*}{129.2} \\
         & min\_samples\_split: [4, 8, 10],     &    &   &   \\
         & n\_estimators: [200, 500]     &    &   &   \\
      \specialrule{0em}{2pt}{2pt}
      \midrule
      \specialrule{0em}{2pt}{2pt}
      \multirow{2}{*}{K-Neighbors} & leaf\_size: [10, 20, 30, 50],   &   \multirow{2}{*}{12}   & \multirow{2}{*}{375.2} & \multirow{2}{*}{5.6} \\
         & n\_neighbors: [3, 5, 7]     &    &   &   \\
      \specialrule{0em}{2pt}{2pt}
      \midrule
      \specialrule{0em}{2pt}{2pt}
      \multirow{2}{*}{Gradient Boosting} & min\_samples\_split: [2, 4, 8, 10],   &   \multirow{2}{*}{16}   & \multirow{2}{*}{394.1} & \multirow{2}{*}{33.3} \\
         & n\_estimators: [100, 200, 300, 500]    &    &   &   \\
      
      \specialrule{0em}{2pt}{2pt}
      \midrule
      \specialrule{0em}{2pt}{2pt}
      \multirow{3}{*}{XGBoost} & learning\_rate: [0.03, 0.1, 0.3],   &   \multirow{3}{*}{27}   & \multirow{3}{*}{373.9} & \multirow{3}{*}{84.7} \\
         & min\_child\_weight: [0.0001, 0.001, 0.01],    &    &   &   \\
         & reg\_lambda: [0.5, 1, 2]    &    &   &   \\
      \specialrule{0em}{2pt}{2pt}
      \midrule
      \specialrule{0em}{2pt}{2pt}
      \multirow{3}{*}{AdaBoost} & learning\_rate: [0.03, 0.3, 1.0],   &   \multirow{3}{*}{9}   & \multirow{3}{*}{881.5} & \multirow{3}{*}{56.3} \\
         & loss: [square, linear],    &    &   &   \\
         & n\_estimators: [50, 100, 500]    &    &   &   \\
      \specialrule{0em}{2pt}{2pt}
      \midrule
      \specialrule{0em}{2pt}{2pt}
      \multirow{3}{*}{MLP} & learning\_rate: [0.0001, 0.001, 0.01],   &   \multirow{3}{*}{27}   & \multirow{3}{*}{351.1} & \multirow{3}{*}{25.9} \\
         & hidden\_layer\_sizes: [(32,),(64,),(32,32)],    &    &   &   \\
         & alpha: [0.001,0.01,0.1]   &    &   &   \\
      \specialrule{0em}{2pt}{2pt}
      \midrule
      \specialrule{0em}{2pt}{2pt}
      \multirow{3}{*}{CatBoost} & learning\_rate: [0.003,0.03,0.1],   &   \multirow{3}{*}{27}   & \multirow{3}{*}{387.4} & \multirow{3}{*}{129.1} \\
         & l2\_leaf\_reg: [0.3, 1.0, 3.0],    &    &   &   \\
         & depth: [4, 7, 10]   &    &   &   \\
      \specialrule{0em}{2pt}{2pt}
   
     \bottomrule
   \end{tabular}}
   \label{Table. diamond learner}
\end{table*}

\end{document}